\documentclass[10pt,twocolumn,letterpaper]{article}

\usepackage{iccv}
\usepackage{times}
\usepackage{epsfig}
\usepackage{graphicx}
\usepackage{amsmath}
\usepackage{amssymb}

\usepackage[T1]{fontenc}
\usepackage{pifont}
\usepackage[innercaption]{sidecap}
\sidecaptionvpos{figure}{t}
\usepackage{enumitem}
\usepackage{booktabs}
\usepackage{amssymb}
\usepackage{amsmath,amsfonts}
\usepackage{amsopn}
\usepackage{bm} 
\usepackage{multirow}
\usepackage{threeparttable}
\usepackage{tabularx}
\usepackage{tablestyles}

\newcommand{\ProbOpr}[1]{\mathbb{#1}}

\newcommand{\expect}[2]{%
\ifthenelse{\equal{#2}{}}{\ProbOpr{E}_{#1}}
{\ifthenelse{\equal{#1}{}}{\ProbOpr{E}\left[#2\right]}{\ProbOpr{E}_{#1}\left[#2\right]}}} 
\newcommand{\var}[2]{%
\ifthenelse{\equal{#2}{}}{\ProbOpr{VAR}_{#1}}
{\ifthenelse{\equal{#1}{}}{\ProbOpr{VAR}\left[#2\right]}{\ProbOpr{VAR}_{#1}\left[#2\right]}}} 

\newcommand{\eat}[1]{}

\usepackage[numbers,sort&compress]{natbib}
\usepackage[pagebackref=true,breaklinks=true,letterpaper=true,colorlinks,bookmarks=false,citecolor=blue]{hyperref}
\newcommand{\cmark}{\ding{51}}%
\newcommand{\xmark}{\ding{55}}%
\newcommand{\ourmethod}{{\sc {MosaicOS}}\xspace}
\newcommand{\ourmethodbf}{{\textbf{\textsc {MosaicOS}}}\xspace}
\newcommand{\ourmethodfull}{\textbf{Mosaic} of \textbf{O}bject-centric images as \textbf{S}cene-centric images\xspace}
\newcommand\mypara[1]{\vspace{0.5mm}\noindent\textbf{#1}}
\newlength\savewidth\newcommand\shline{\noalign{\global\savewidth\arrayrulewidth\global\arrayrulewidth1pt}\hline\noalign{\global\arrayrulewidth\savewidth}}
\definecolor{peach}{rgb}{1.0, 0.8, 0.64}

\usepackage[breaklinks=true,bookmarks=false]{hyperref}

\iccvfinalcopy 

\begin{document}

\title{\textbf{\ourmethodbf:} A Simple and Effective Use of Object-Centric Images\\ for Long-Tailed Object Detection}

\author{Cheng Zhang$^{1}$\thanks{Equal contributions} \quad Tai-Yu Pan$^1$\footnotemark[1] \quad Yandong Li$^2$ \quad Hexiang Hu$^3$ \\
Dong Xuan$^1$ \quad Soravit Changpinyo$^2$ \quad Boqing Gong$^2$ \quad Wei-Lun Chao$^1$ \\[10pt]
{$^1$The Ohio State University \quad  $^2$Google Research \quad $^3$University of Southern California \quad }\\
}

\maketitle
\ificcvfinal\thispagestyle{empty}\fi

\begin{abstract}
Many objects do not appear frequently enough in complex scenes (\eg, certain handbags in living rooms) for training an accurate object detector, but are often found frequently by themselves (\eg, in product images). Yet, these \emph{object-centric} images are not effectively leveraged for improving object detection in \emph{scene-centric} images. In this paper, we propose \textbf{Mosaic} of \textbf{O}bject-centric images as \textbf{S}cene-centric images (\ourmethod), a simple and novel framework that is surprisingly effective at tackling the challenges of long-tailed object detection.
Keys to our approach are three-fold: (i) pseudo scene-centric image construction from object-centric images for mitigating domain differences, (ii) high-quality bounding box imputation using the object-centric images' class labels, and (iii) a multi-stage training procedure. On LVIS object detection (and instance segmentation), \ourmethod leads to a massive $60\%$ (and $23\%$) relative improvement in average precision for rare object categories.
We also show that our framework can be compatibly used with other existing approaches to achieve even further gains.
Our pre-trained models are publicly available at \url{https://github.com/czhang0528/MosaicOS/}.
\end{abstract}

\section{Introduction}
\label{s_intro}
 
Detecting objects in complex daily scenes is a long-standing task in computer vision \cite{girshick2014rich,viola2001robust,papageorgiou1998general,felzenszwalb2009object}. 
With rapid advances in deep neural networks \cite{he2016deep,simonyan2015very,szegedy2015going,krizhevsky2012imagenet,huang2017densely} and the emergence of large-scale datasets \cite{lin2014microsoft,kuznetsova2020open,shao2019objects365,yu2020bdd100k,thomee2015yfcc100m}, there has been remarkable progress in detecting \emph{common} objects (\eg, cars, humans, \etc)~\cite{he2017mask,ren2015faster,liu2016ssd,redmon2016you,lin2017focal,lin2017feature,bochkovskiy2020yolov4,zoph2020rethinking}.
However, detecting \emph{rare} objects (\eg, unicycles, bird feeders, \etc) proves much more challenging due to the inherent limitation of training data. In particular, complex scenes in which an object appears pose another variation factor that is too diverse to capture from a small amount of data~\cite{gupta2019lvis,inat,zhu2014capturing}.
Developing algorithms to overcome such a ``long-tailed'' distribution of object instances in \emph{scene-centric} images (SCI) \cite{gupta2019lvis,lin2014microsoft,shao2019objects365,kuznetsova2020open} has thus attracted a flurry of research interests \cite{tan2020equalization,wang2020frustratingly,li2020overcoming}.

\begin{figure}[t]
    \centerline{\includegraphics[width=0.92\linewidth]{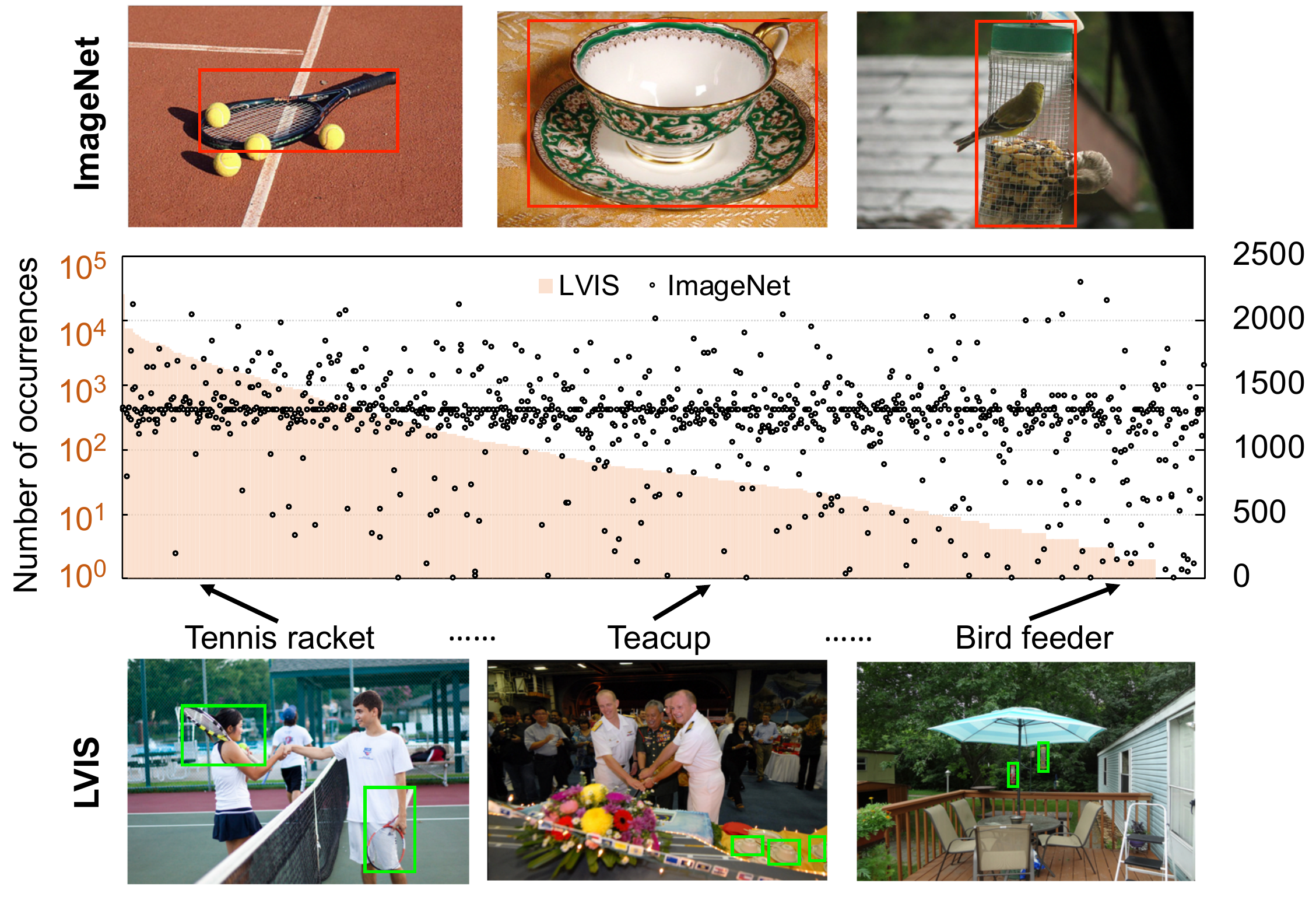}}
    \vskip -2pt
    \caption{\small \textbf{Object frequencies in scene-centric and object-centric images.} \textbf{\color{peach}Orange} bars show the number of instances per class in the scene-centric LVIS v0.5 dataset~\cite{gupta2019lvis}. Class indices are sorted by the instance numbers. Black dots show  the number of images  in the object-centric ImageNet datasets~\cite{deng2009imagenet} for each corresponding class. \emph{The two types of images have very different trends of object frequencies.} We also show three examples of both datasets, corresponding to frequent, common, and rare classes in LVIS. \textbf{\color{red}{Red}} and \textbf{\color{green}{green}} boxes indicate the objects. \emph{These two types of images have different focuses and object sizes.}}
    \label{fig:1}
    \vskip -10pt
\end{figure}

Fortunately, the uncommon objects in scene-centric images often appear more frequently in \emph{object-centric} images (OCI) in which the objects of interest occupy the main and most salient regions (\eg, product images).
For example, given ``bird feeder'' as query, a popular image search engine (\eg, Google Images) mostly retrieves object-centric ``bird feeder'' results. Similarly, curated object recognition datasets such as ImageNet~\cite{deng2009imagenet} contain more than a thousand object-centric ``bird feeder'' images, nearly a hundred times more than scene-centric images from LVIS v0.5~\cite{gupta2019lvis}. We further illustrate this point in \autoref{fig:1}, in which a discrepancy in frequencies of the same objects in ImageNet and LVIS can generally be observed (see \S~\ref{s_data} for details).

Can we leverage such abundant object-centric images to improve long-tailed object detection? 
The most common approach to this is to leverage these images for pre-training the object detector's backbone~\cite{zoph2020rethinking,he2019rethinking,he2017mask}. While this general approach may benefit various tasks beyond object detection, it is highly data-intensive and does not take care of the domain gap between the pre-training and downstream tasks (see \S~\ref{ss_further} for analysis). As a result, they do not always improve the object detection accuracy \cite{he2019rethinking,zoph2020rethinking}.

In this paper, we propose \ourmethod (\ourmethodfull), a simple and effective framework to leverage object-centric images for object detection.
\ourmethod directly uses object-centric images during the \emph{training} of object detectors.
There are three key ingredients. The first one is the construction of \emph{pseudo} scene-centric images from object-centric images using mosaic\footnote{Mosaic was exploited in~\cite{bochkovskiy2020yolov4,chen2020stitcher,zhou2020cheaper}, but mainly to combine multiple \emph{scene-centric} images to simulate smaller object sizes or to increase the scene complexity, not to turn object-centric images into scene-centric ones.}. The second one is the imputation of bounding box annotations using image class labels. 
The third ingredient is a multi-stage training procedure for learning from both gold scene-centric and synthesized pseudo scene-centric annotations. 
\autoref{fig:framework} illustrates our framework.

Our use of mosaic and bounding box imputation to construct \emph{pseudo} scene-centric images from object-centric images tackles two key challenges in leveraging object-centric images for object detection.
The first challenge is a ``domain" gap between object-centric and scene-centric images: an object-centric image usually contains fewer (but bigger) object instances and a less complex background and this discrepancy is believed to unfavorably hinder knowledge transfer \cite{zamir2018taskonomy,wang2019characterizing}.
The second challenge is missing detection annotations: object-centric images, either from the Internet or object recognition datasets (\eg, ImageNet), are not \emph{perfectly object-centric}, usually provided \emph{without} accurate object localization in the form of bounding boxes.

Our proposed framework leads to significant accuracy gains in long-tailed object detection and instance segmentation on LVIS \cite{gupta2019lvis}, using object-centric images from ImageNet \cite{deng2009imagenet} and the Internet.
In particular, for the task of object detection for rare objects, we observe a significant boost from $13\%$ to over $20\%$ in average precision.
For the task of instance segmentation, our approach even improves the accuracy on common objects. More importantly, unlike the baseline approaches, we do so without sacrificing the accuracy on the frequent classes.
Finally, we also explore combining our approach with existing techniques~\cite{ren2020balanced} that results in even better performance.

\mypara{Our main contributions} are summarized as follows:
\begin{itemize}[leftmargin=*,topsep=0pt,itemsep=0pt,noitemsep]
    \item Bringing the best of object-centric images to the long-tailed object detection on scene-centric images. 
    \item Algorithms for mosaicking and pseudo-labeling to mitigate the domain discrepancy between two image types.
    \item A multi-stage training framework that leverages the pseudo scene-centric images (from object-centric images) to improve the detector on scene-centric images.
    \item Extensive evaluation and analysis of the proposed approach on the challenging LVIS benchmark~\cite{gupta2019lvis}.
\end{itemize}
\section{Related Work}
\label{s_related}

\mypara{Long-tailed object detection}
has attracted increasing attentions recently. The challenge is the drastically low accuracy for detecting rare objects.
Most existing works develop training strategies or objectives to address this \cite{gupta2019lvis,hu2020learning,li2020overcoming,ramanathan2020dlwl,wang2020frustratingly,tan2020equalization,wang2020devil,tang2020long,ren2020balanced,wu2020forest}.
Wang et al.~\cite{wang2020devil} found that the major performance drop is by mis-classification, suggesting the applicability of class-imbalanced classification methods (\eg, re-weighting, re-sampling)~\cite{krishna2017visual,guo2016ms,lin2014microsoft,thomee2015yfcc100m,inat,finetunemajorfeat,cao2019learning,CBfocal,BasicSMOTE2002,BasicLearnfromImb,CBfocal,buda2018systematic,kang2019decoupling,BasicLearnfromImb}.
Different from them, we study an alternative and orthogonal solution to the problem (\ie, exploiting abundant object-centric images).

\mypara{Weakly-supervised or semi-supervised object detection}
learns or improves object detectors using images with weak supervision (\eg, image-level labels) \cite{li2016weakly,gao2019note,uijlings2018revisiting,bilen2016weakly,divvala2014learning} or even without supervision \cite{jeong2019consistency,li2020improving,gao2019note,sohn2020simple,zoph2020rethinking}. They either leverage scene-centric images or detect only a small number of common classes (\eg, classes in Pascal VOC \cite{everingham2010pascal}, MSCOCO \cite{lin2014microsoft}, or ILSVRC \cite{russakovsky2015imagenet}).
Our work can be seen as weakly supervised object detection, but we focus on the challenging long-tailed detection with more than $1,000$ objects. Meanwhile, we leverage object-centric images, which is different from scene-centric images in both appearances and layouts. The most related work is \cite{ramanathan2020dlwl}, which uses the YFCC-100M dataset \cite{thomee2015yfcc100m} (Flickr images) to improve the detection on LVIS \cite{gupta2019lvis}. However, YFCC-100M contains both object-centric images and scene-centric images and a non-negligible label noises. Thus, \cite{ramanathan2020dlwl} employs more sophisticated data pre-processing and pseudo-labeling steps, yet our approach achieves higher accuracy (see \autoref{table:data}). 

Other works use object-centric images to expand the label space of the object detector \cite{kuen2019scaling,hoffman2014lsda,tang2017visual,tang2016large,hoffman2015detector,hoffman2016large,redmon2017yolo9000}.
Such approaches mostly only use object-centric images to learn the last fully-connected classification layer, instead of improving the features extractor. In contrast, our approach can improve the feature extractor, and successfully transfer knowledge to long-tailed instance segmentation.
\section{Scene-Centric vs. Object-Centric Images}
\label{s_data}

Images taken by humans can roughly be categorized into object-centric and scene-centric images. The former captures objects of interest (\eg, cats) and usually contains just one salient class whose name is used as the image label. The later captures a scene and usually contains multiple object instances of different classes in a complex background.

Recent object detection methods mainly focus on scene-centric images \cite{lin2014microsoft,gupta2019lvis,shao2019objects365}. Since scene-centric images are not intended to capture specific objects, \emph{object frequencies in our daily lives will likely be reflected in the images.} As such, the learned detector will have a hard time detecting rare objects: it just has not seen sufficient instances to understand the objects' appearances, shapes, variations, etc.

In contrast, humans tend to take object-centric pictures that capture interesting (and likely uncommon, rare) objects, especially during events or activities (\eg, bird watching, \etc). Thus, a rare object in our daily lives may occur more often in the online object-centric images.

\mypara{Discrepancy \wrt object frequencies.}
We compare object frequencies of the ImageNet \cite{deng2009imagenet} and LVIS (v0.5) \cite{gupta2019lvis} datasets. The former 
retrieved images from the Internet by querying search engines using the object class names (thus object-centric). Whereas the later used MSCOCO \cite{lin2014microsoft} dataset, which collects daily scene images with many common objects in a natural context (thus scene-centric).

The full ImageNet has $21,841$ classes, whereas LVIS has around $1,230$ classes. Using the WordNet synsets \cite{miller1995wordnet},
we can match $1,025$ classes ($1,016$ classes are downloadable) between them. \autoref{fig:1} shows the number of object instances per class in LVIS and the number of images per corresponding class in ImageNet.
It presents \emph{a huge difference between object frequencies of these two datasets.} For example, ImageNet has a balanced distribution across classes and LVIS is extremely long-tailed. Even for rare classes in LVIS (those with $<10$ training images), ImageNet usually contains more than $1,000$ images. Such a difference offers an opportunity to resolve the long-tailed object detection in scene-centric images via the help of object-centric images.

\mypara{Discrepancy \wrt visual appearances and contents.}
Beyond frequencies, these two types of images also have other, less favorable discrepancies. The obvious one is the number of object instances per image. LVIS on average has \textbf{12.1} labeled object instances per image (the median number of instances per image is \textbf{6}).
While most of the ImageNet images are not annotated with object bounding boxes, according to a subset of images used in the ILSVRC detection challenge \cite{russakovsky2015imagenet}, each image has \textbf{2.8} object instances. The larger number of object instances, together with the intention behinds the images, implies that scene-centric images also have \emph{smaller} objects in size and more complex backgrounds. This type of discrepancies, contrast to that in object frequencies, is not favorable for leveraging the object-centric images, and may lead to negative transfer \cite{zamir2018taskonomy,wang2019characterizing}.
\section{Overall Framework}
\label{s_framework}

\begin{figure*}[t]
    \centerline{\includegraphics[width=1\linewidth]{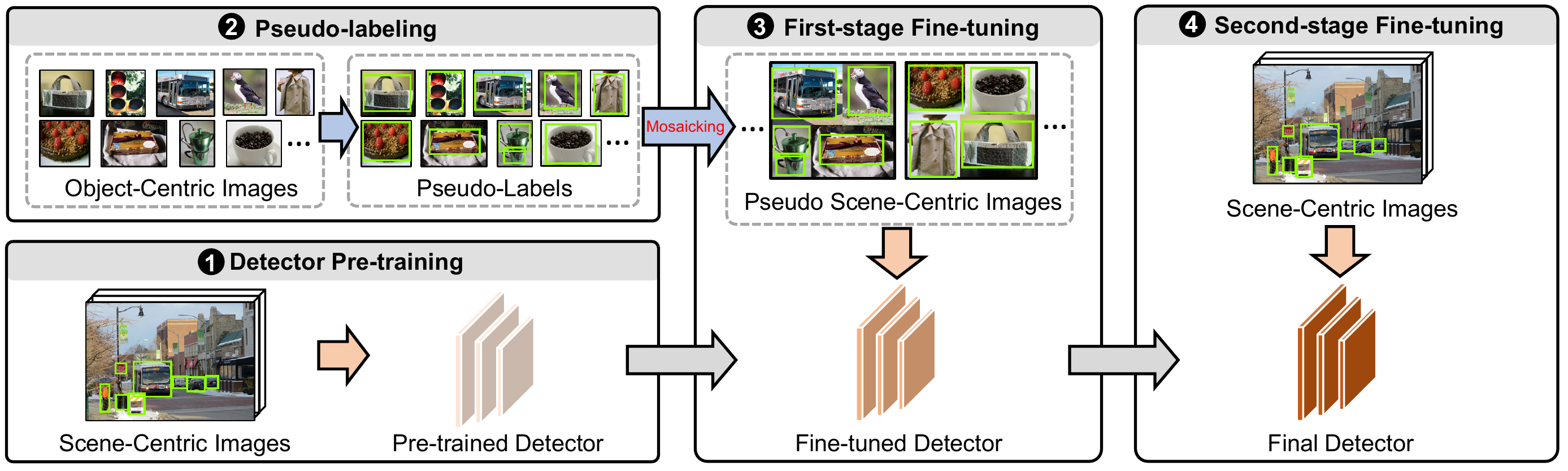}}
    \caption{\small \textbf{Our \ourmethod framework for leveraging object-centric images for long-tailed object detection.} It consists of four stages. \textbf{\ding{182} Detector Pretraining}: we pre-train an object detector using scene-centric images with gold-labeled box annotations. \textbf{\ding{183} Pseudo-labeling}: we construct pseudo scene-centric images from object-centric images using box annotation imputation (possibly using the pre-trained detector in stage $1$) as well as mosaicking (stitching multiple images together). \textbf{\ding{184} First-stage Fine-tuning}: we fine-tune the pre-trained detector from stage $1$ with pseudo scene-centric images from stage $2$. \textbf{\ding{185} Second-stage Fine-tuning}: we further fine-tune the object detector from stage $3$ using scene-centric images with gold-labeled box annotations again, similar to stage $1$. {\color{orange}{Orange}} arrows indicate data feeding for training. {\color{gray}Gray} arrows indicate model cloning. {\color{green}Green} boxes indicate the (pseudo \& gold-labeled) box annotations.}
    \label{fig:framework}
    \vskip -10pt
\end{figure*}

To better leverage object-centric images for object detection, we present a novel learning framework, which includes \textbf{three simple\footnote{We claim our approach to be ``simple" as it employs simple methods to address the fundamental challenges. Pseudo-labeling is an essential step to use weakly-supervised data, and we apply simple fixed locations.
We apply mosaicking and multi-stage training to bridge the domain gap, instead of applying sophisticated methods like domain adversarial training \cite{ganin2016domain}.} yet effective components} to handle
\begin{enumerate} [label=(\alph*), noitemsep,topsep=2pt]
    \item the domain gap between two image sources;
    \item the missing bounding box labels;
    \item the integration of both image sources for training.
\end{enumerate}
\autoref{fig:framework} gives an illustration of our framework. Concretely, the framework begins with pre-training an object detector using the accurately labeled scene-centric images. Any object detector can be applied. Without loss of generality, we focus on Faster R-CNN \cite{ren2015faster}, one of the most popular object detectors in the literature. The pre-trained object detector serves for two purposes: it can help impute the missing boxes in object-centric images; it will be used as the initialization for training with the object-centric images.

To turn object-centric images into training examples for object detection, we must handle both (a) and (b). \emph{We postpone the details of these two components to \S~\ref{s_approach}.} For now, let us assume that we have processed and labeled object-centric images with pseudo ground-truth boxes and class labels like the labeled scene-centric images.
To differentiate from the original object-centric images, we call the new images \emph{pseudo scene-centric images} (see \autoref{fig:framework}).

The pseudo scene-centric images may still have domain gaps from real scene-centric images, to which the learned detector will finally be applied. Besides, the pseudo ground-truth boxes may contain noises (\eg, wrong locations). To effectively learn from these images (especially for rare objects) while not sacrificing the detector's ultimate accuracy on identifying and locating objects,
we propose to \emph{fine-tune
the pre-trained object detector via two stages.}
In what follows, we first give a brief review of object detection.

\mypara{Backgrounds on object detection.} An object detector has to identify objects with their class names and locate each of them by a bounding box. Taking Faster R-CNN \cite{ren2015faster} as an example, it first generates a set of object proposals (usually around $512$) that likely contain objects of interest. This is done by the region proposal network (RPN) \cite{ren2015faster}. Faster R-CNN then goes through each proposal to identify its class (can be ``background'') and refine the box location and size.

The entire Faster R-CNN is learned with three loss terms
\begin{align}
    \mathcal{L} = \mathcal{L}_\text{rpn} + \mathcal{L}_\text{cls} + \mathcal{L}_\text{reg},
    \label{eq:1}
\end{align}
where $\mathcal{L}_\text{rpn}$ is for RPN training, $\mathcal{L}_\text{cls}$ is for multi-class classification, and $\mathcal{L}_\text{reg}$ is for box refinement.

\mypara{Two-stage fine-tuning.} 
Given the pre-trained detector, we first fine-tune it using the pseudo scene-centric images that are generated from object-centric images (see~\S~\ref{s_approach}). We then fine-tune it using the labeled scene-centric images. We separate the two image sources since they are still different in appearances and label qualities. The second stage helps adapt the detector back to real scene-centric images.

In both stages, all the three loss terms in \autoref{eq:1} are optimized. {We do not freeze any parameters except the batch-norm layers~\cite{ioffe2015batch} in the backbone feature network which are kept frozen by default.} We will compare to single-stage fine-tuning with both images and fine-tuning using only $\mathcal{L}_\text{cls}$ for pseudo scene-centric images in \S~\ref{s_exp}.

\section{Creating Pseudo Scene-Centric Images}
\label{s_approach}

We now focus on the missing components of our framework: generating pseudo scene-centric images from object-centric images. Our goal is to create images that are more \emph{scene-centric-like} and label them with pseudo ground truths.
We collect object-centric images from two sources: ImageNet \cite{deng2009imagenet} and Google Images.
See \S~\ref{ss_setup} for details.

\subsection{Assigning Pseudo-Labels}
\label{ss_plabel}
Each object-centric image has one object class label, but no bounding box annotations. Some images may contain multiple object instances and classes, in which the class label only indicates the most salient object. Our goal here is to create a set of pseudo ground-truth bounding boxes that likely contain the object instances for each image, and assign each of them a class label, such that we can use the image to directly fine-tune an object detector.

\begin{figure}[t]
    \centerline{\includegraphics[width=1\linewidth]{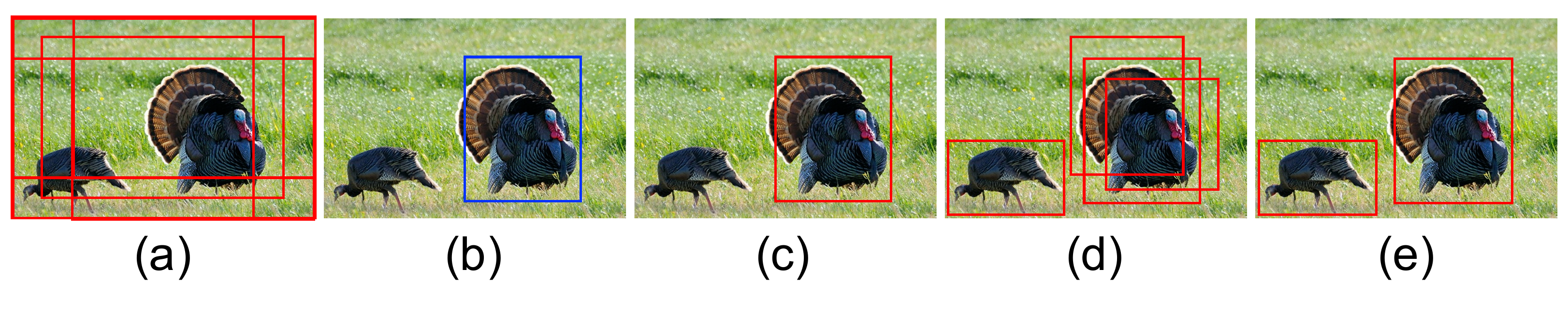}}
    \vskip-5pt
     \caption{\small \textbf{A comparison of pseudo-label generation:} (a) fixed locations, (b) trust the detector, (c) trust the detector + image labels, (d) trust the calibrated detector + image labels, and (e) localization by region removal.
     The image label is ``turkey,'' a rare class in LVIS. {\color{red}{Red}}/{\color{blue}{blue}}  boxes are labeled as ``turkey''/other classes.}
    \label{fig:pseudo-comp}
    \vskip -10pt
\end{figure}

There are indeed many works on doing so, especially those for weakly-supervised and semi-supervised object detection~\cite{li2016weakly,uijlings2018revisiting,bilen2016weakly,divvala2014learning,li2020improving,gao2019note,ramanathan2020dlwl}. {The purpose of this subsection is therefore not to propose a new way to compare with them, but to investigate approaches that are more effective and efficient in a large-scale long-tailed setting.}
Specifically, we investigate five methods that do not require an extra detector or proposal network beyond the pre-trained one. \emph{As will be seen in~\S~\ref{ss_exp_det}, imputing the box class labels using the image class label is the key to success.} \autoref{fig:pseudo-comp} illustrates the difference of these methods. {Please see the {\color{magenta}supplementary material} for other possibilities.}

\mypara{Fixed locations (F).} We simply assign some fixed locations of an object-centric image to be the pseudo ground-truth boxes, regardless of the image content. The hypothesis is that many of the object-centric images may just focus on one object instance whose location is likely in the centre of the image (\ie, just like those in \cite{griffin2007caltech,fei2004learning}). Specifically, we investigate the combination of the whole image, the center crop, and the four corner crops: in total \textbf{six} boxes per image. The height and width of the crops are $80\%$ of the original image. We assign each box the image class label.

\mypara{Trust the pre-trained detector (D).} We apply the pre-trained detector learned with the scene-centric images to the object-centric images, and treat the detected boxes and predicted class labels as the pseudo-labels. Specifically, we keep all the detection of confidence scores $>0.5$\footnote{$0.5$ is the default threshold for visualizing the detection results.}. We apply non-maximum suppression (NMS) among the detected boxes of each class using an IoU (intersection-over-union) threshold $0.5$. By doing so, every image will have boxes of different sizes and locations, labeled with different classes.

\mypara{Trust the pre-trained detector \& image class labels (D$\dagger$).} One drawback of the above method is its tendency to assign high-frequency labels, a notorious problem in class-imbalanced learning \cite{ye2020identifying,kang2019decoupling,tan2020equalization}. For instance, if ``bird'' is a frequent class and ``eagle'' is a rare class, the detector may correctly locate an ``eagle'' in the image but assign the label ``bird'' to it. To resolve this issue, we choose to trust the box locations generated by the above method but assign each box the image class label instead of the predicted class labels. In other words, a box initially labeled as ``bird'' is replaced by the label ``eagle'' if ``eagle'' is the image label. The rationale is that in an object-centric image, most of the object instances belong to the image's class. 

\mypara{Trust the calibrated detector \& image class labels (D$\ddagger$).} Another way to resolve the above issue is to set for each class a different confidence threshold\footnote{For Faster R-CNN, each RPN proposal can lead to multiple detected boxes, one for each class whose probability is larger than the threshold.}. The rational is that a classifier trained with long-tailed data tends to assign lower probabilities to minor classes of scarce training data~\cite{Kim2020Adjusting,kang2019decoupling,ye2020identifying,buda2018systematic}. We thus reduce the threshold for each class according to its number of training images. Let $N_\text{max}$ be the size of the most major class and let $N_\text{c}$ be the size of class $c$, we apply a threshold $0.5 \times (N_\text{c}/N_\text{max})^\gamma$ for class $c$, inspired by \cite{mahajan2018exploring}. We set $\gamma=0.5$ according to validation. Compared to the vanilla ``trust the detector,'' this method will generate more boxes for common and rare classes. We again replace their detected labels by the image class label\footnote{Without doing so, the approach can hardly improve ``trust the detector'' due to more noisy boxes being included. See the {\color{magenta}supplementary} for details.}.

\mypara{Localization by region removal (LORE).}
We investigate yet another way for pseudo-labeling, taking the following intuition: an image classifier should fail to predict the correct label if the true object regions are removed. To this end, we first train a ResNet-50 \cite{he2009learning} image classifier using our object-centric image pool\footnote{That is, we train the classifier with these images, and then apply this classifier back to these images (after some regions are removed).}.
We then collect the pre-trained detector's predicted boxes on these images, trusting the box locations but not the class labels. We sort these boxes by how much removing each boxed region alone reduces the image classifier's confidence on the image label. 
We then remove these boxed regions \emph{in turn} until the classifier fails to predict the true label. The bounding boxes of the removed regions are then collected as the pseudo ground truths for the image. We assign each box the image class label. Please see the {\color{magenta}supplementary material} for details.

\subsection{Synthesizing Pseudo Scene-centric Images}
\label{ss_synthesizing}
We apply a simple technique, \ie, \emph{image mosaic}, to make object-centric images more scene-centric, in terms of appearances, contents, and layouts. Concretely, we stitch multiple object-centric images together to obtain a new image that contains more object instances and classes, smaller object sizes, and more complex background. Specifically, we stitch $2\times 2$ images together, which are sampled either randomly within a class, or randomly from the entire image pool.
We do not apply sophisticated stitching tools like \cite{brown2007automatic,zhang2014parallax,hays2007scene} but simply concatenate these images one-by-one. The resulting images are thus more like \emph{mosaics}, having artifacts along the stitched boundaries 
(see \autoref{fig:framework}).

\section{Experiments}
\label{s_exp}

We conduct experiments and analysis for \ourmethod, on the tasks of long-tailed Object Detection (OD) and Instance Segmentation (IS). We begin by introducing the experimental setup (\S~\ref{ss_setup}), then present the main object detection results as well as detailed ablation studies (\S~\ref{ss_exp_det}, \S~\ref{ss_further}), and finally show additional results that evaluate our model on instance segmentation and the other dataset (\S~\ref{ss_exp_seg}). \emph{We include qualitative results in the {\color{magenta}supplementary material.}}

\subsection{Setup}
\label{ss_setup}
\mypara{Long-tailed OD \& IS datasets and metrics.}
We evaluate our approach on LVIS instance segmentation benchmark \cite{gupta2019lvis}. We focus on v0.5 as most existing works, and report additional key results on v1.0 (more in the {\color{magenta}supplementary}).
LVIS v0.5 contains $1,230$ entry-level object categories with around 2 million high-quality annotations. 
The training set contains all the classes with a total of $57,623$ image; the validation set contains $830$ classes with a total of $5,000$ images. The categories are naturally long-tailed distributed and are divided into three groups based on the number of training images per class: rare (1-10 images), common (11-100 images), and frequent ($>$100 images).
\emph{All results are reported on the validation set.}
We adopt the standard mean average precision (AP) metric in LVIS \cite{gupta2019lvis}. \emph{We specifically focus on object detection using the standard bounding box evaluation, $\text{AP}^{b}$.} The AP on the rare, common, and frequent classes ($\text{AP}_{r}^{b}$, $\text{AP}_{c}^{b}$, $\text{AP}_{f}^{b}$) are also reported separately. 

\mypara{Object-centric data sources.}
We mainly use images from two sources: ImageNet~\cite{deng2009imagenet} and Google Images~\cite{google}. ImageNet is a classification benchmark. Most people use its 
$1,000$ categories version in ILSVRC \cite{russakovsky2015imagenet} and treat it as the standard dataset for backbone pre-training in various computer vision tasks. 
The full version of ImageNet has $21,842$ classes. In LVIS and ImageNet, each category has a unique WordNet synset ID \cite{miller1995wordnet}, and we are able to match $1,016$ LVIS classes and retrieve the corresponding images from ImageNet (in total, $769,238$ images).
Besides, we retrieve images via Google by querying with class names and descriptions provided by LVIS. Such a text-to-image search returns hundreds of iconic images and we take the top $100$ for each of the $1,230$ classes.

\mypara{Implementation.}
We use {Faster R-CNN} \cite{ren2015faster}\footnote{Our implementation is based on \cite{wu2019detectron2}, which uses RoIAlign \cite{he2017mask} instead of RoIPool~\cite{girshick2015fast} to extract region features for R-CNN training.} as our base detector and ResNet-50 \cite{he2016deep} with a Feature Pyramid Network (FPN) \cite{lin2017feature} as the backbone, which is pre-trained on ImageNet (ILSVRC) \cite{russakovsky2015imagenet}. Our base detector is trained on the LVIS training set with \emph{repeated factor sampling}, following the standard training procedure in \cite{gupta2019lvis} (1x schedule). To fairly compare with our fine-tuning results, we further extend the training process with another 90K iterations and select the checkpoint with the best $\text{AP}^{b}$ as \textbf{Faster R-CNN$\star$}. The following experiments are initialized by {Faster R-CNN$\star$}. See the {\color{magenta}supplementary} for more details.

\mypara{Baselines.}
We compare to the following baselines: 
\begin{itemize} [noitemsep,topsep=2pt,leftmargin=*]
    \item \textbf{Self-training} is a strong baseline for semi-supervised learning \cite{lee2013pseudo}. 
    We follow the state-of-the-art self-training method for detection \cite{zoph2020rethinking} and use {Faster R-CNN$\star$} to create pseudo-labels on the object-centric images, same as ``trust the pre-trained detector''. We then fine-tune {Faster R-CNN$\star$} using both pseudo scene-centric and LVIS images for 90K iterations, with the normalized loss~\cite{zoph2020rethinking}.
    \item \textbf{Single-stage fine-tuning} fine-tunes {Faster R-CNN$\star$} with both pseudo scene-centric and LVIS images in one stage. In each mini-batch, we have 50\% of data from each source. Different ratios do not lead to notable differences.
    \item \textbf{DLWL}~\cite{ramanathan2020dlwl} is the state-of-the-art method that uses extra unlabeled images from the YFCC-100M~\cite{thomee2015yfcc100m}.
\end{itemize}
For self-training and single-stage training, we perform $2\times2$ mosaicking to create pseudo scene-centric images.

\mypara{Variants of \ourmethod.} (a) We compare different object-centric image sources and their combinations. 
(b) We compare with or without mosaicking.
(c) For mosaicking, we compare stitching images from the same classes (so the artifacts can be reduced) or from randomly selected images. 
(d) We compare different ways to generate pseudo-labels (see~\S~\ref{ss_plabel}). (e) We study fine-tuning with pseudo scene-centric images using only the classification loss $\mathcal{L}_\text{cls}$.

\begin{table*}[ht]
\tabcolsep 9.5pt
\renewcommand\arraystretch{1.0}
\small
\centering
\caption{\small \textbf{Comparison of object detection on LVIS v0.5 validation set.} \textbf{OCIs:}  object-centric images sources (IN: ImageNet, G: Google). \textbf{Mosaic:} \cmark means 2$\times$2 image mosaicking. \textbf{Hybrid:} \cmark means stitching images from different classes. \textbf{P-GT:} ways to generate pseudo-labels (\textbf{F}: six \text{f}ixed locations, \textbf{D}: trust the  \text{d}etector, \textbf{D$\dagger$}: trust the  \text{d}etector and image \text{l}abel, \textbf{D$\ddagger$}: trust the calibrated detector and image class label, \textbf{L} (LORE): \text{l}ocalization by region removal, and \textbf{S}: a {s}ingle box of the whole image). The best result per column is in bold.}
\begin{tabular}{l|cccc|cccccc}
\multicolumn{1}{c|}{} & OCIs & Mosaic & Hybrid  & P-GT & $\text{AP}^{b}$ & $\text{AP}^{b}_{50}$  & $\text{AP}^{b}_{75}$  & $\text{AP}^{b}_{r}$   & $\text{AP}^{b}_{c}$   & $\text{AP}^{b}_{f}$ \\
\shline
Faster R-CNN & - & - & - & -  & 23.17 & 38.94 & 24.06 & 12.64 & 22.40 & 28.33 \\
Faster R-CNN$\texttt{$\star$}$ & - & - & - & - & 23.35 & 39.15 & 24.15 & 12.98 & 22.60 & 28.42 \\
 \hline
\multirow{2}{*}{\begin{tabular}[c]{@{}c@{}}{Self-training~\cite{zoph2020rethinking}}\end{tabular}}  & IN & \cmark & \cmark  & D
& 22.71 & 38.22 & 23.79 & 14.52 & 21.41 & 27.61 \\
 & IN & \cmark & \cmark  & F & 23.46 & 39.03 & 24.82 & 16.20 & 22.19 & 27.94 \\
 \hline
Single-stage & IN & \cmark & \cmark  & F & 20.09 & 35.34  & 20.27  & 12.96  & 19.08  & 24.20  \\
 \hline
\multirow{9}{50pt}{{\ourmethod (Two-stage)}} & IN & {\color{red}{\xmark}} & \xmark  & F & 24.27 & 40.30 & 25.61 & 16.97 & 23.29 & 28.42 \\

 & IN & {\color{red}\cmark} & {\color{teal}{\xmark}}  &F & 24.48 & 40.12 & 25.65 & 18.76 & 23.26 & 28.29 \\
 \cline{2-11}
 & IN & \cmark & \cmark   & {\color{blue}D} & 23.04 & 38.97 & 23.72 & 13.93 & 21.51 & 28.14 \\
  & IN & \cmark & \cmark  & ~~{\color{blue}D$\dagger$} & 24.66 & 40.31 & 25.99 & 17.45 & 23.62 & {28.83} \\
& IN & \cmark & \cmark  & ~~{\color{blue}D$\ddagger$} & 24.93 & 40.48 &  26.71 & 19.31 & 23.51 & \textbf{28.95} \\
  & IN & \cmark & \cmark  & {\color{blue}S} & 24.59 & 40.20  & 25.78  &  19.13 &  23.35 & 28.32 \\
 
 & {IN} & \cmark & \cmark  & {\color{blue}L} & 24.83 & 40.58 & 26.27 & 20.06 & 23.25 & {28.71} \\
 
& {\color{brown}IN} & \cmark & {\color{teal}\cmark}  & {\color{blue}F} & 24.75 & 40.44 & 26.09 & 19.73 & 23.44 & {28.39} \\
 \cline{2-11}
 & {\color{brown}IN+G} & \cmark & \cmark  & F & \textbf{25.01} & \textbf{40.76} & \textbf{26.46} & \textbf{20.25} & \textbf{23.89} & 28.32 \\
 \hline
\end{tabular}\label{table:main}  \vspace*{-10pt}
\end{table*}

\subsection{Results on Object Detection}
\label{ss_exp_det}
\mypara{Main results.}  \autoref{table:main} summarizes the results. 
It shows that the model trained with pseudo-labels generated by six fixed location (F) has a very competitive performance comparing to other strategies. We therefore consider it as the default pseudo-labeling method given its simplicity and effectiveness.
Meanwhile, our two-stage approach with object-centric images outperforms Faster R-CNN$\star$ (and Faster R-CNN) notably. On $\text{AP}_{r}^{b}$ for rare classes, our best result of $20.25\%$ is $\sim7.2\%$ higher than Faster R-CNN$\star$, justifying our motivation: {object-centric images that are resistant to the long-tailed nature of object frequencies can improve object detection in scene-centric images.}

\mypara{Mosaicking is useful}~{\color{red}(red in~\autoref{table:main})}. A simple $2\times 2$ stitching leads to a notable gain: $\sim1.8\%$ at $\text{AP}_{r}^{b}$, supporting our claim that making object-centric images similar to scene-centric images is important. 
Indeed, according to~\S~\ref{s_data}, a $2\times 2$ stitched image will have around $12$ objects, very close to that in LVIS images. Stitching images from different classes further leads to a $\sim1.0\%$ gain {\color{teal}(green in~\autoref{table:main})}.

\mypara{Fixed-location boxes are effective}~{\color{blue}(blue in~\autoref{table:main})}. By comparing different ways for pseudo-labels, we found that both localization by region removal (L) and the simple six fixed locations (F) lead to 
strong results without querying the pre-trained detector.
Using six fixed locations slightly outperforms using one location (S) (\ie, the image boundary), probably due to the effect of data augmentation.
All the three methods significantly outperform ``trust the pre-trained detector'' (D), and we attribute this to the poor pre-trained detector's accuracy on rare classes: it either cannot identify rare classes from object-centric images or is biased to detect frequent classes. \textbf{By replacing the detected classes with the image labels} and/or further calibrating the detector for more detected boxes, \ie, ``trust the (calibrated) pre-trained detector and image label'' ({D$\dagger$}, {D$\ddagger$}), we see a notable boost, which supports our claim. Nevertheless, they are either on par with or worse than fixed locations (F), especially on $\text{AP}_{r}^{b}$ for rare classes, again showing the surprising efficacy of the simple method. We note that, both {D$\dagger$} and {D$\ddagger$} are specifically designed in this work for long-tailed problems and should not be seen as existing baselines.

\begin{SCfigure}[]
  {\includegraphics[width=0.45\linewidth]{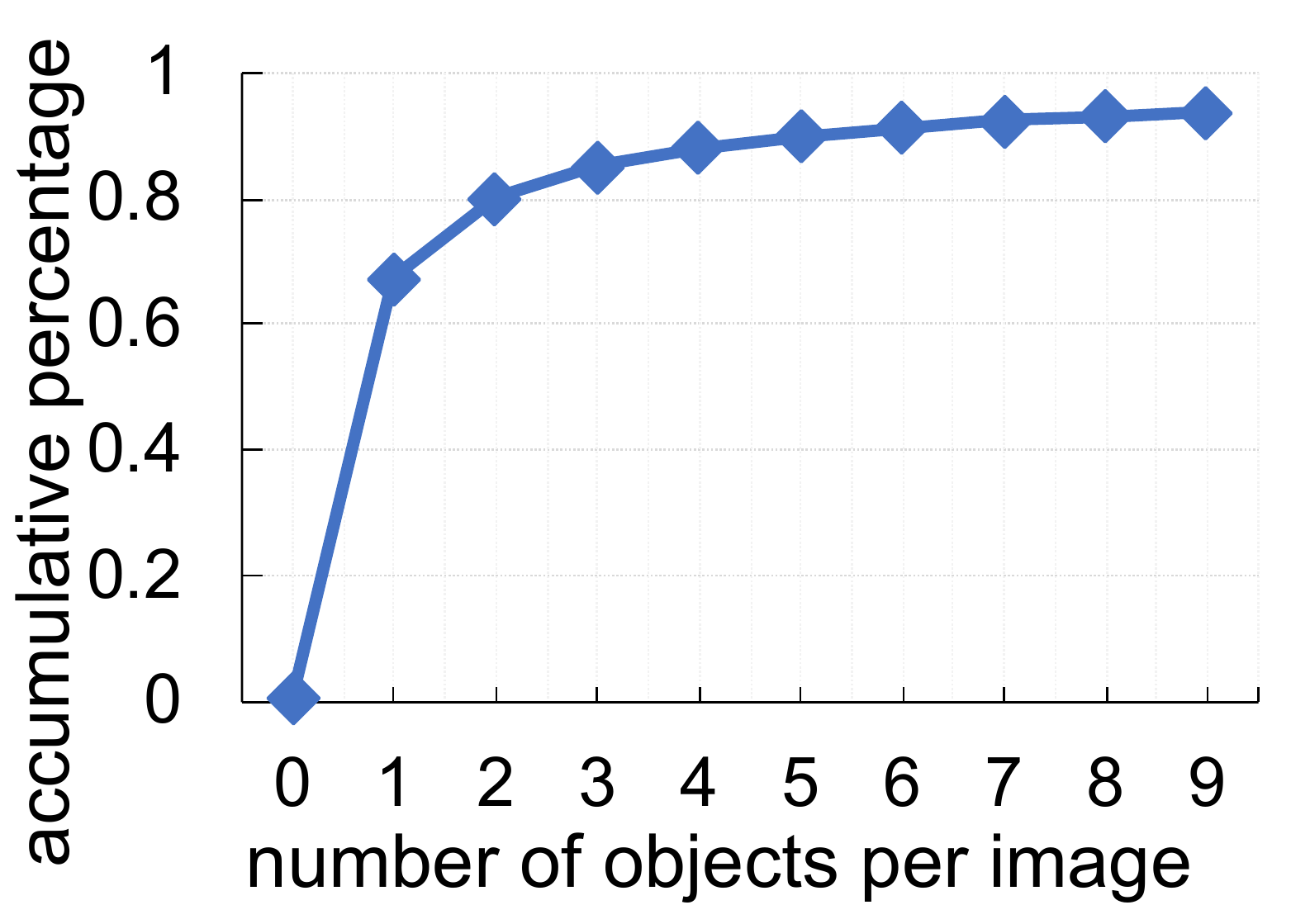}\vspace{-15pt}}  
  \caption{\small \textbf{\# of objects per object-centric image found by LORE.} We use ImageNet images. Y-axis is the accumulative percentage of images whose objects are no more than the X-axis number.\vspace{-15pt}}
  \label{fig:number_object}
\end{SCfigure}

To further analyze why fixed locations work well, we check the numbers of boxes LORE finds per image. LORE keeps removing regions until the classifier fails to classify the image. The number of regions it found is thus an estimation of the number of target objects (those of the image label) in the image. \autoref{fig:number_object} shows the accumulative number of images whose object numbers are no more than a threshold: $\sim70\%$ of ImageNet images have one target object instance, suggesting that it may not be necessary to locate and separate object instances in pseudo-labeling.

\mypara{Self-training and single-stage fine-tuning.}
As shown in \autoref{table:main}, self-training (with D and loss normalization~\cite{zoph2020rethinking})
outperforms Faster R-CNN\texttt{$\star$} on $\text{AP}^{b}_{r}$. As self-training is sensitive to the pseudo-label quality, we also apply the fixed locations (F) to it and achieve improvement. By comparing it to single-stage fine-tuning (with F), we see the benefit of loss normalization between the two image sources.

By comparing self-training to its counterparts in two-stage fine-tuning (with D and F), we however find that two-stage fine-tuning leads to higher accuracy in most cases.
This demonstrates the benefit of separating image sources in fine-tuning, in which the second stage adapts the detector back to accurately labeled true scene-centric images. 

\mypara{The amount of object-centric data}~{\color{brown}(brown in~\autoref{table:main})}. As ImageNet only covers $1,016$ classes of LVIS, we augment it with $100$ Google images per class for all the $1,230$ LVIS classes. We see another $0.5\%$ gain at rare classes ($\text{AP}_{r}^{b}$).

\emph{For the following analyses besides \autoref{table:main}, we will focus on our approach with two-stage fine-tuning, ImageNet object-centric images, $2\times 2$ mosaic with images from multiple classes, and fixed locations (F) as the pseudo-labels.}

\mypara{Losses in fine-tuning.} 
We compare using all three losses of Faster R-CNN (\ie, $\mathcal{L}_\text{rpn}+\mathcal{L}_\text{cls}+\mathcal{L}_\text{reg}$) or just the classification loss (\ie, $\mathcal{L}_\text{cls}$) in the first-stage fine-tuning with pseudo scene-centric images. \autoref{table:classification_loss} shows the results. The former outperforms the latter on three out of four metrics. This tells that, while the pseudo boxes do not accurately bound the objects, learning the RPN and box refinement with them (\eg, to predict a high objectness score) is still beneficial.

\begin{table}[t]
\small
\tabcolsep 9pt
\renewcommand\arraystretch{1.0}
\centering
\caption{\small \textbf{Losses in the first fine-tuning stage.}}
\begin{tabular}{l|cccc}
Losses & $\text{AP}^{b}$ & $\text{AP}_{r}^{b}$   & $\text{AP}_{c}^{b}$   & $\text{AP}_{f}^{b}$  \\
\shline
$\mathcal{L}_\text{cls}$ & 24.53  & 18.87 & 23.07 & 28.61 \\
$\mathcal{L}_\text{rpn}+\mathcal{L}_\text{cls}+\mathcal{L}_\text{reg}$ & 24.75 & 19.73 & 23.44 & 28.39 \\
\hline
\end{tabular}\label{table:classification_loss}
\vskip -12pt
\end{table}

\mypara{Other baselines.} We compare to state-of-the-art methods that use no extra object-centric images in~\autoref{table:sota_detection}.  We obtain comparable or better results, especially on rare classes.

\mypara{Compatibility with existing efforts.} Our approach is compatible with recent efforts in better backbone pre-training \cite{li2020overcoming} and advanced training objectives (\eg, \cite{ren2020balanced}). For instance, following BaGS~\cite{li2020overcoming} to pre-train the backbone using MSCOCO images, we achieve an improved $26.28$ $\text{AP}^b$ (see \autoref{table:sota_detection}). Further incorporating the balanced loss \cite{ren2020balanced} into the second-stage fine-tuning boosts $\text{AP}^b$ to $28.06$. 

\begin{table}[t]
\small
\tabcolsep 3.8pt
\renewcommand\arraystretch{1.0}
\centering
\caption{\small {{\textbf{Object detection on LVIS v0.5}. We use ImageNet + Google Images. MSCOCO: for pre-training. \cite{ren2020balanced}: balanced loss.}}}
\begin{tabular}{r|cc|cccc}
\multicolumn{1}{c|}{} & MSCOCO &\cite{ren2020balanced} & $\text{AP}^{b}$ & $\text{AP}_{r}^{b}$   & $\text{AP}_{c}^{b}$   & $\text{AP}_{f}^{b}$  \\
\shline
BaGS~\cite{li2020overcoming} & \cmark & & 25.96 & 17.65 & 25.75 & 29.54 \\
TFA~\cite{wang2020frustratingly} & &&24.40 & 16.90 & 24.30 & 27.70 \\
\hline
\multirow{4}{*}{\ourmethod} &&& 25.01 & {20.25} & 23.89 & 28.32 \\
&\cmark&& 26.28 & 17.37 & 26.13 & 30.02 \\
& &\cmark&  26.83 & \textbf{21.00} & 26.31 & 29.81\\
&\cmark&\cmark&  \textbf{28.06} & 19.11 & \textbf{28.23} & \textbf{31.41}\\
\hline
\end{tabular}\label{table:sota_detection}
\vskip -5pt
\end{table}

\subsection{Detailed Analysis of \ourmethodbf}
\label{ss_further}

\mypara{The importance of object-centric images.} In \autoref{table:main}, we show that even without mosaic, the use of object-centric images already notably improves the baseline ($\text{AP}^{b}$: $24.27$ vs. $23.35$).
We further investigate the importance of mosaic of object-centric images: our use of mosaic is different from \cite{bochkovskiy2020yolov4,chen2020stitcher,zhou2020cheaper}, which stitch scene-centric images in the training set to simulate smaller objects or increase the scene complexity. We apply their methods to stitch LVIS images and study two variants: stitching scene-centric images~\cite{bochkovskiy2020yolov4,chen2020stitcher} or the cropped objects~\cite{zhou2020cheaper} from them. \autoref{table:suppl_stitching} shows that \ourmethod surpasses both variants on rare and common objects, justifying the importance of incorporating ample object-centric images to capture the diverse appearances of objects, especially for rare objects 
in scene-centric images.

\begin{table}[t]
\tabcolsep 5pt
\renewcommand\arraystretch{1.0}
\centering
\small
\caption{\small {\textbf{The importance of mosaicking object-centric images.} SCI: object-centric images in the original LVIS training set.} }
\begin{tabular}{l|cccc}
         & $\text{AP}^{b}$  & $\text{AP}_{r}^{b}$ & $\text{AP}_{c}^{b}$ & $\text{AP}_{f}^{b}$ \\
        \shline
        Faster R-CNN$\star$ & 23.35 & 12.98 & 22.60 & 28.42 \\
        \hline
        Stitching SCI~\cite{bochkovskiy2020yolov4} & 23.83 & 13.99 & 23.02 & \textbf{28.76} \\
        Stitching SCI~\cite{chen2020stitcher} & 23.58 & 14.00 & 22.58 & {28.66} \\
        Stitching cropped SCI~\cite{zhou2020cheaper} & 23.55 & 13.40 & 23.04 & 28.26\\
        \hline
        \ourmethod & \textbf{24.75} & \textbf{19.73} & \textbf{23.44} & {28.39} \\
        \hline
        \end{tabular}\label{table:suppl_stitching}
\vskip-10pt
\end{table}

\begin{figure}[t]
    \centerline{\includegraphics[width=1\linewidth]{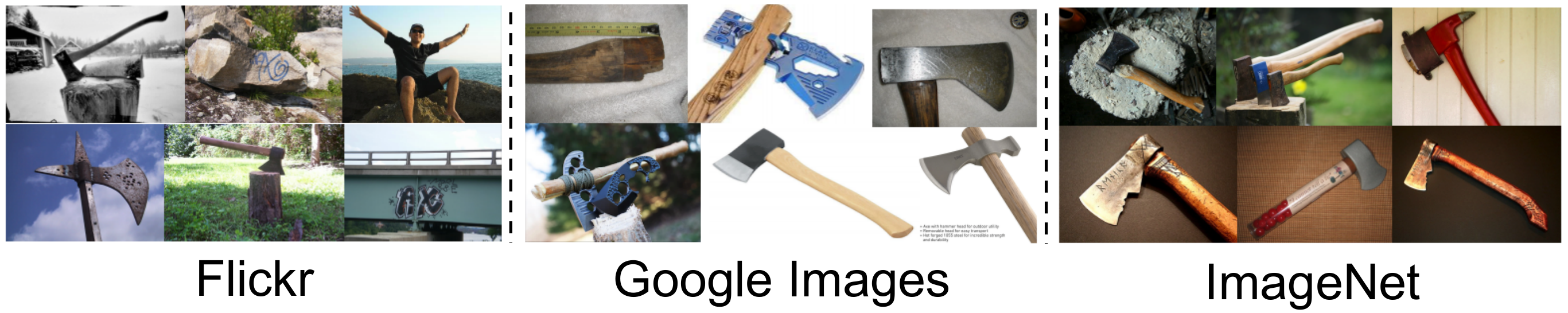}}
    \vskip -5pt
     \caption{\small \textbf{A comparison of object-centric image sources.} We show images of a rare class (``ax'') in LVIS. ImageNet~\cite{deng2009imagenet} (right) and Google Images~\cite{google} (middle) give images with more salient ``ax'' inside, while Flickr~\cite{flickr} (left) gives more noisy images, either with very small axes, cluttered backgrounds, or even no axes.}
    \label{fig:data_source}
    \vskip -5pt
\end{figure}

\mypara{Does the quality of data sources matter?} DLWL \cite{ramanathan2020dlwl} uses YFCC-100M \cite{thomee2015yfcc100m}, a much larger data source than ImageNet. YFCC-100M images are mainly collected from Flickr, which mixes object-centric and scene-centric images and contains higher label noises. DLWL \cite{ramanathan2020dlwl} thus develops sophisticated pre-processing and pseudo-labeling steps. In contrast, we specifically leverage \emph{object-centric} images that have higher object frequencies and usually contain only single object classes (the image labels), leading to a much simpler approach.
As shown in \autoref{table:data}, our method (with IN) outperforms DLWL by a large margin: $>5.5\%$ at $\text{AP}_{r}^{b}$. We attribute this to our ways of strategically collecting object-centric images and stitching them to make them scene-centric-like. The fact that we identify a better data source should not lead to an impression that we merely solve a simpler problem, but an evidence that selecting the right data source is crucial to simplify a problem. \autoref{fig:data_source} illustrates the difference among these sources. 

For a fair comparison to \cite{ramanathan2020dlwl} in terms of the algorithms, we also investigate Flickr images.
Since~\cite{ramanathan2020dlwl} does not provide their processed data, we directly crawl Flickr images (100 per class) and re-train our algorithm. We achieve $24.05/16.17$ $\text{AP}^{b}/\text{AP}^{b}_r$, better than DLWL. Using pure Google images beyond ImageNet can achieve $24.45/19.09$. 
Our novelties and contributions thus lie in both the algorithm and the direction we investigate. The latter specifically leads to simpler solutions but higher accuracy. 

\begin{table}[t]
\small
\tabcolsep 4pt
\renewcommand\arraystretch{1.0}
\centering
\caption{\small {{\textbf{Comparison of object detection on LVIS v0.5 using different extra data sources.}} G: Google Images. IN: ImageNet.}}
\begin{tabular}{l|l|cccc}
\multicolumn{1}{c|}{} & \multicolumn{1}{c|}{Data} & $\text{AP}^{b}$ & $\text{AP}_{r}^{b}$   & $\text{AP}_{c}^{b}$   & $\text{AP}_{f}^{b}$  \\
\shline
Faster R-CNN$\texttt{$\star$}$ & -- & 23.35 & 12.98 & 22.60 &  28.42 \\
\hline
DLWL~\cite{ramanathan2020dlwl} & YFCC-100M & 22.14 & 14.21 & - &  -\\
\hline
\multirow{3}{*}{\ourmethod} & Flickr & 24.05 & 16.17 & 23.06 & 28.43  \\
& G & 24.45 & 19.09 & 23.27 & 28.08 \\
& IN &  24.75 & {19.73} & 23.44 & 28.39 \\
\hline
\end{tabular}\label{table:data}
\vskip -5pt
\end{table}

\begin{table}[]
\small

\tabcolsep 9.5pt
\renewcommand\arraystretch{1.0}
\centering
\caption{\small {\textbf{Object detection on the 176 overlapped classes} between ImageNet-1K (ILSVRC) and LVIS v0.5.}}
\begin{tabular}{l|cccc}
 & $\text{AP}^{b}$   & $\text{AP}_{r}^{b}$   & $\text{AP}_{c}^{b}$   & $\text{AP}_{f}^{b}$ \\ 
\shline
\# Category & 176 & 21 & 84 & 71 \\
\hline
Faster R-CNN$\texttt{$\star$}$ & 26.05 & 14.78 & 23.92 & 31.16 \\
\ourmethod & \textbf{27.50} & \textbf{21.16}  & \textbf{25.45} & \textbf{31.80}  \\
\hline
\end{tabular}\label{table:imgnet_overlap}
\vskip -10pt
\end{table}

\mypara{The importance of learning for the downstream tasks.}
We found $176$ classes of LVIS validation set in ILSVRC. That is, the corresponding ImageNet images used by \ourmethod are already seen by the pre-trained detector's backbone. Surprisingly, as shown in \autoref{table:imgnet_overlap}, \ourmethod still leads to a notable gain for these classes, which not only justifies its efficacy, but also suggests the importance of learning the downstream tasks directly with those images.

\subsection{Results on Instance Segmentation \& LVIS v1.0}
\label{ss_exp_seg}

\mypara{Instance segmentation.}
We also validate our approach on instance segmentation, in a similar manner: we prepare pseudo scene-centric images \emph{with box labels} and use them in the first fine-tuning stage by optimizing the losses in \autoref{eq:1}. That is, we do not optimize segmentation losses in this stage.
We apply Mask R-CNN~\cite{he2017mask} with ResNet-50 as the backbone. \autoref{table:sota_segmentation} shows the results: \emph{the baselines do not use extra object-centric images}. We see a notable gain against vanilla Mask R-CNN for rare and common classes, even if we have no segmentation labels on object-centric images. This supports the claim in \cite{wang2020devil}: even for detection and segmentation, the long-tailed problem is mainly in the classification sub-network. We perform on par with the state-of-the-art methods. Details are in the {\color{magenta}supplementary}.

\begin{table}[t]
\tabcolsep 8.5pt
\renewcommand\arraystretch{1.0}
\centering
\small
\caption{\small {\textbf{Instance segmentation on LVIS v0.5.}
We use images from IN + G as \autoref{table:main}. +~\cite{ren2020balanced}: include the balanced loss.}}
\begin{tabular}{r|cccc}
\multicolumn{1}{c|}{} & $\text{AP}$ & $\text{AP}_{r}$   & $\text{AP}_{c}$   & $\text{AP}_{f}$  \\
\shline
Mask R-CNN~\cite{gupta2019lvis} & 24.38 & 15.98 & 23.96 & 28.27 \\
BaGS~\cite{li2020overcoming} & 26.25 & 17.97 & 26.91 & 28.74\\
BALMS~\cite{ren2020balanced} & 27.00 & 19.60 & \textbf{28.90} & 27.50 \\
\hline
\ourmethod & 26.26 & 19.63 & 26.60 & 28.49 \\
\ourmethod+~\cite{ren2020balanced} &  \textbf{27.86} & \textbf{20.44} & 28.82 & \textbf{29.62}\\
\hline
\end{tabular}\label{table:sota_segmentation}
\vskip -5pt
\end{table}

\mypara{LVIS v1.0 results.} 
We highlight consistent empirical results on LVIS v1.0, where our approach wins in both object detection and instance segmentation, using ResNet-50-FPN (see \autoref{table:lvis_v1_det}). More comparisons are in the {\color{magenta}supplementary}.

\begin{table}[]
\small
\tabcolsep 4.2pt
\renewcommand\arraystretch{1.0}
\centering
\caption{\small \textbf{LVIS v1.0 Results.} We report both object detection and instance segmentation performances of our method.}
\begin{tabular}{l|cccccc}
{OD Results} & $\text{AP}^{b}$    &$\text{AP}_{50}^{b}$  &$\text{AP}_{75}^{b}$ &$\text{AP}_{r}^{b}$   & $\text{AP}_{c}^{b}$   & $\text{AP}_{f}^{b}$ \\ 
\shline
Faster R-CNN$\star$ & 22.01&	36.36&	23.14&	10.57 & 20.09 & 29.18 \\ 
\ourmethod & \textbf{23.90}  & \textbf{38.61} &\textbf{25.32} & \textbf{15.45} &	\textbf{22.39} &	\textbf{29.30} \\ 
\hline
{IS Results} & $\text{AP}$ & $\text{AP}_{50}$  &$\text{AP}_{75}$ &$\text{AP}_{r}$   & $\text{AP}_{c}$   & $\text{AP}_{f}$ \\ 
\shline
Mask R-CNN & 22.59 &	35.44&	23.87&	12.31 & 21.30 & 28.55 \\
\ourmethod &\textbf{24.49}  & \textbf{38.02} &\textbf{25.87} & \textbf{18.30} &	\textbf{23.00} &	\textbf{28.87} \\ 
										
\hline
\end{tabular}
\label{table:lvis_v1_det}
\vskip -10pt
\end{table}
\section{Discussion and Conclusion}
\label{s_disc}
We investigate the use of object-centric images to facilitate long-tailed object detection on scene-centric images. We propose a concrete framework for this idea that is both simple and effective. Our results are encouraging, improving the baseline by a large margin on not only detecting but also segmenting rare objects. We hope that our study can attract more attention in using these already available but less explored object-centric images to overcome the long-tailed problem. 
Please see the {\color{magenta}supplementary} for more discussion.

{\small \mypara{Acknowledgments.} We are thankful for the generous support by Ohio Supercomputer Center and AWS Cloud Credits for Research.}

{\small
\bibliographystyle{ieee_fullname}
\bibliography{main}
}

\clearpage
\appendix
\begin{center}
\section*{Supplementary Material}
\end{center}
In this Supplementary Material, we provide details and results omitted in the main text.
\begin{itemize}
    \item \autoref{suppl_contribution}: contributions. (\S~\ref{s_disc} of the main paper)
    \item \autoref{suppl_pseudo_labels}: additional details and results on pseudo-label generation. (\S~\ref{ss_plabel} of the main paper)
    \item \autoref{suppl_mosaic}: ablation studies on image mosaicking. (\S~\ref{ss_synthesizing} of the main paper)
    \item \autoref{suppl_self_train}: further analysis on self-training. (\S~\ref{ss_setup} and \S~\ref{ss_exp_det} of the main paper)
    \item \autoref{suppl_oci}: further analysis on data quality of object-centric images. (\S~\ref{s_data} and \S~\ref{ss_further} of the main paper)
    \item \autoref{suppl_adv_training}: comparison to adversarial training. (\S~\ref{s_framework} of the main paper)
	\item \autoref{suppl_impl}: implementation details of \ourmethod on object detection and instance segmentation. (\S~\ref{ss_setup} and \S~\ref{ss_exp_seg} of the main paper)
	\item \autoref{suppl_v0.5}: detailed results on LVIS v0.5. (\S~\ref{ss_exp_det} and \S~\ref{ss_exp_seg} of the main paper)
	\item \autoref{suppl_lvis_v1}: detailed results on LVIS v1.0. (\S~\ref{ss_exp_seg} of the main paper)
    \item \autoref{suppl_qual}: qualitative results of object detection on LVIS v0.5. (\S~\ref{ss_exp_det} of the main paper).
\end{itemize}

\section{Contribution and Novelty}
\label{suppl_contribution}

Our main contributions are in the idea of using object-centric images (OCI) to facilitate long-tailed object detection on scene-centric images (SCI) as well as a concrete implementation of this idea that is both simple and effective. This is by no means trivial; for instance, a related work~\cite{ramanathan2020dlwl} with a more sophisticated approach can hardly improve the accuracy (\autoref{table:data} of the main paper). While most existing works focus on \emph{designing new algorithms} to learn from long-tailed data, our proposal is orthogonal to them, and can be combined together for further improvement.

Although leveraging auxiliary data to improve \emph{common} object detection has been studied previously, existing works typically assume access to well prepared data from a similar visual domain, with sufficient object instances.
However, collecting and annotating such auxiliary data is extremely challenging in \emph{long-tailed} object detection. In contrast, our method does not have such a limitation as we make use of \emph{object-centric} images readily available over the Internet (via search engines), which contains sufficient object instances though in a slight different domain.
Particularly, we observe that making use of such rich \emph{object-centric} images (from ImageNet) leads to more superior empirical performances against \cite{ramanathan2020dlwl}, which uses YFCC-100M \cite{thomee2015yfcc100m}.

To enable more general applicability, we make the design of our framework as straightforward as possible. Along this process, two challenges are identified, \ie, the gap between visual domains and the lack of object labels. To address them, we investigate simple algorithms such as fixed box locations, mosaicking, and multi-stage training.
We note that more sophisticated techniques 
can be incorporated as well. The facts that (a) \emph{our framework performs on par with state-of-the-art long-tailed detection methods} and (b) \emph{many existing techniques can be easily plugged into our framework} further justify the potential of this promising direction.

\begin{figure}[t]
    \centerline{\includegraphics[width=1.0\linewidth]{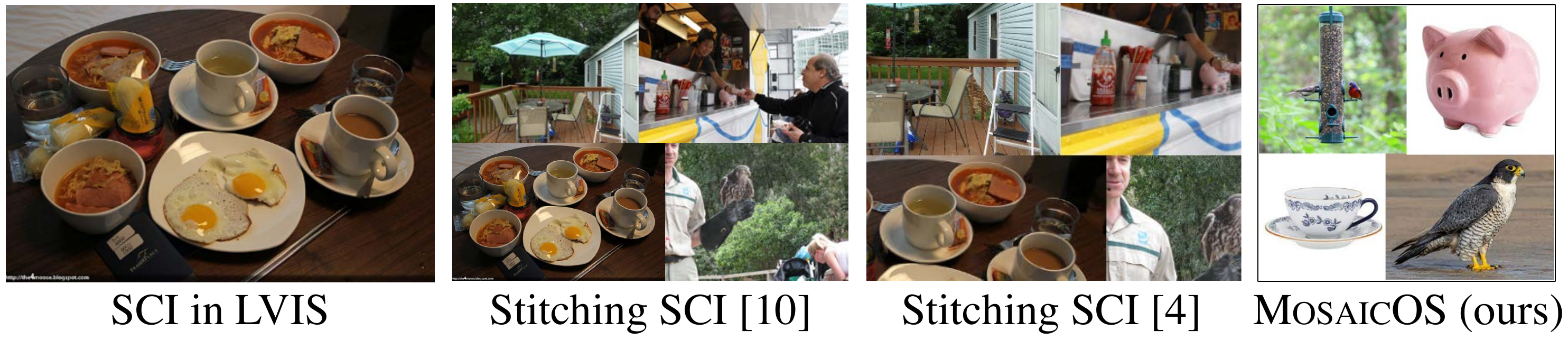}}
    \caption{\small \textbf{Different stitching methods.} \ourmethod introduces \emph{more diverse} examples by leveraging object-centric images, while existing methods~\cite{bochkovskiy2020yolov4,chen2020stitcher} only perform data augmentation using scene-centric images.}
    \vskip -10pt
    \label{fig:suppl_stitching}
\end{figure}

While several components of our framework --- mosaic, pseudo-labeling, two-stage fine-tuning --- have been individually explored in prior works in different contexts, \emph{a suitable combination is essential and novel} for our idea to work. 
Further, our use of mosaic on OCI is different from \cite{bochkovskiy2020yolov4,chen2020stitcher}, as shown in \autoref{fig:suppl_stitching}.
Our contributions also include extensive analysis that justifies the importance of each component.
These insights led to a simple and effective framework, which we consider a strength. For example, our LORE approach (\S~\ref{suppl_lore}) could have provided methodological novelty. But its small gain over simple fixed locations does not justify the inclusion of it into our final framework.

\section{Pseudo-Label Generation}
\label{suppl_pseudo_labels}

\subsection{Trust the calibrated detector and image labels}
We provide analysis on pseudo-label generation with detector calibration and imputation using image class labels.

\begin{figure}[t]
    \centerline{\includegraphics[width=1\linewidth]{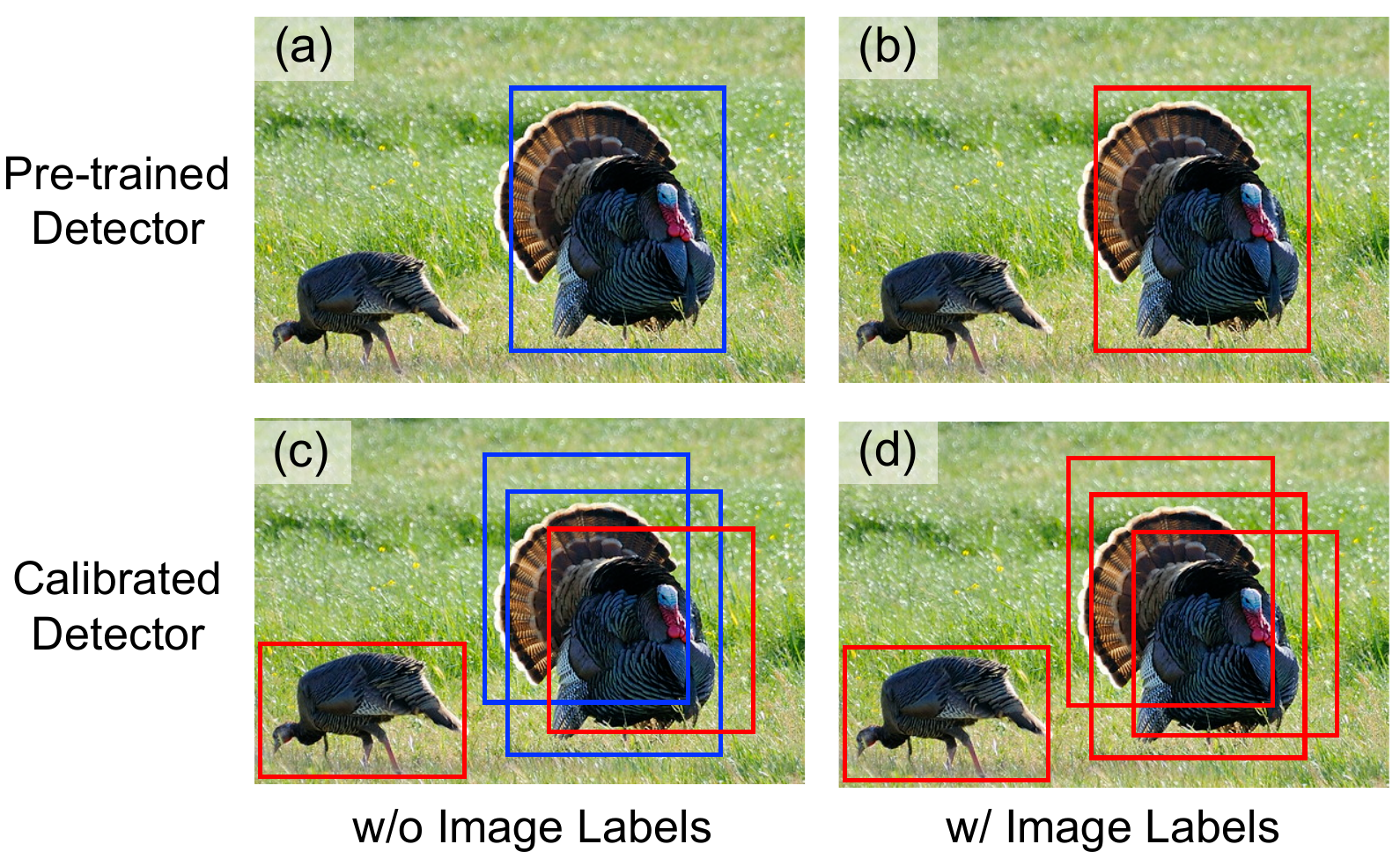}}
    \caption{\small \textbf{A comparison of pseudo-label generation with detector calibration and imputation using image class labels.} (a) trust the detector (D), (b) trust the detector + image class labels (D$\dagger$), (c) trust the calibrated detector, and (d) trust the calibrated detector + class image labels (D$\ddagger$). The image label is ``turkey'', a rare class in LVIS. {\color{red}{Red}}/{\color{blue}{Blue}} boxes are labeled as ``turkey''/other classes. See \S~\ref{ss_plabel} of the main paper for details.}
    \label{fig:suppl_calib_detector}
\end{figure}

\mypara{Detector calibration.}
As mentioned in \S~\ref{ss_plabel} of the main paper, we calibrate the pre-trained detector by assigning each class a different confidence threshold according to the class size --- rare classes have lower thresholds.
\autoref{fig:suppl_calib_detector} illustrates the difference with and without detector calibration, and with and without imputation using the image class labels. By assigning each class a different confidence threshold, the calibrated detector outputs more detected boxes, indicating that many rare and common objects are missed by the pre-trained detector due to low confidence scores (\autoref{fig:suppl_calib_detector} (a) vs. (c)). However, simply applying calibration can hardly correct the wrong labels that have already been biased toward the frequent classes (blue boxes in \autoref{fig:suppl_calib_detector} (c)). Next, we explore the idea of bringing the best of image class labels to correct noisy detected labels.

\mypara{The importance of imputation with image class labels.}
For object-centric images, most of the object instances belong to the image’s class label. We therefore improve ``trust the pre-trained detector'' (\autoref{fig:suppl_calib_detector} (a)) and ``trust the calibrated detector'' (\autoref{fig:suppl_calib_detector} (c)) by assigning each box the image class label (see \autoref{fig:suppl_calib_detector} (b) and (d)). As shown in \autoref{fig:suppl_calib_detector} and \autoref{table:calib_detector}, we see significant improvements for both the pre-trained and calibrated detectors. Specifically, assigning the image class label for each box can largely boost the performance for rare objects ($\text{AP}^{b}_{r}$).

\subsection{Details on LORE}
\label{suppl_lore}

\begin{figure*}[t]
    \centerline{\includegraphics[width=0.9\linewidth]{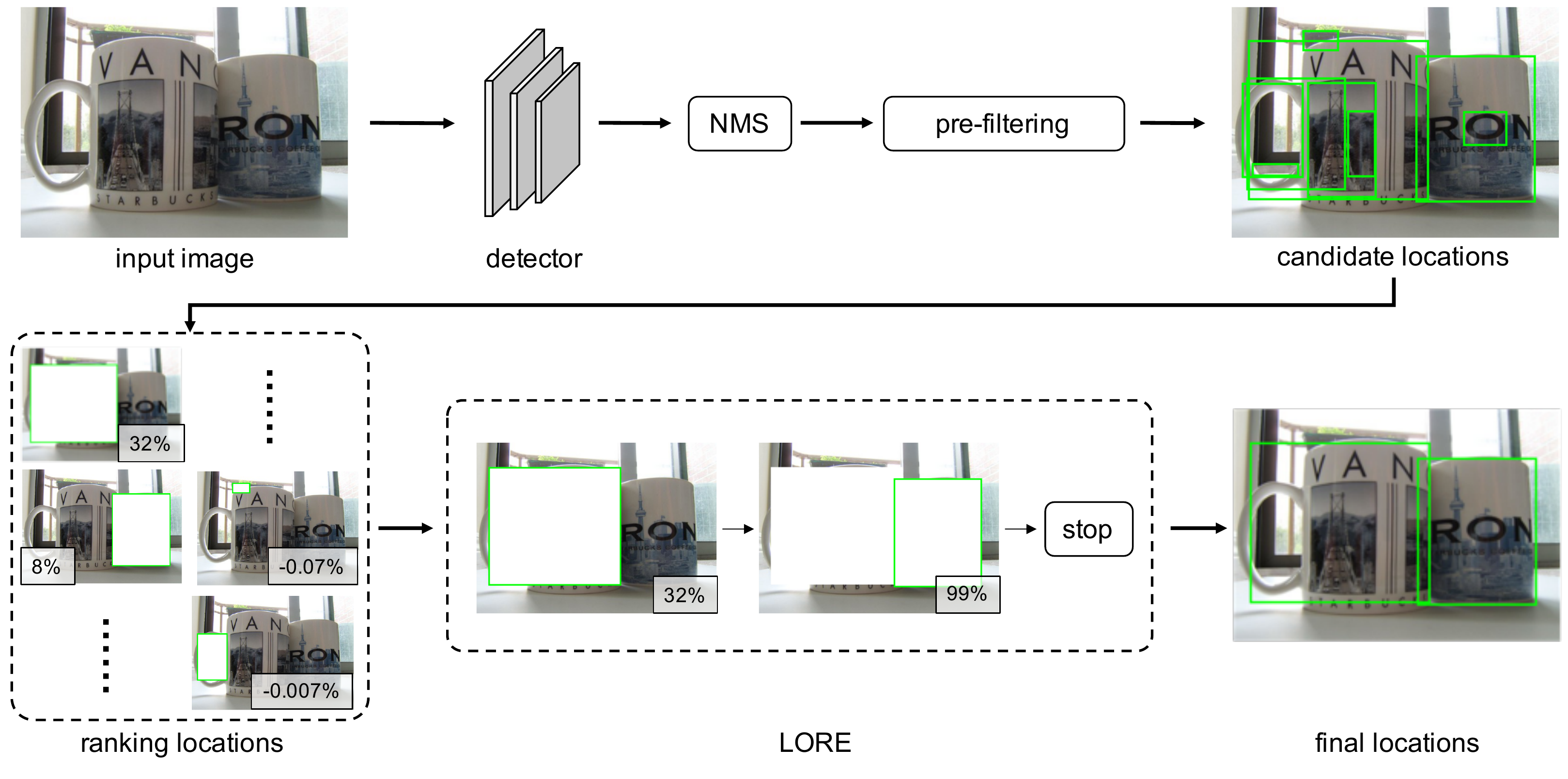}}
    \caption{\small \textbf{Illustration of LORE.} We first apply a pre-trained detector to obtain candidate boxes, followed by pre-filtering. We then sort the remaining boxes using an image classifier. Finally, we remove the boxes in turn until the classifier fail to predict the target image label. The numbers at image corners indicate the confidence reducing ratio. Negative values mean the confidence increases after removing outliers.}
    \label{fig:suppl_lore2}
\end{figure*}

\autoref{fig:suppl_lore2} shows the pipeline of localization by region removal (LORE), which is introduced in \S~\ref{ss_plabel} of the main paper for pseudo-label generation. Concretely, LORE takes an object-centric image as the input and identifies the locations of the target object (\ie, that of image label) in the image. The whole pipeline consists of three major components: (1) classifier training, (2) box pre-filtering, and (3) localization by removal. We describe each step as follows.

\mypara{Classifier training.}
We train a ResNet-50~\cite{he2016deep} image classifier with all object-centric images. For LVIS v0.5 dataset, we follow the conventional training procedure\footnote{\url{https://github.com/pytorch/examples/tree/master/imagenet}} to train a $1,230$-way ResNet classifier. Specifically, we train the networks with 90 epoch and achieve 74$\%$ top-1 training accuracy. We use this pre-trained classifier to rank object regions in object-centric images. 

\mypara{Box pre-filtering.}
We feed an object-centric image into the pre-trained \emph{object detector} and collect detection results. Concretely, we take the top 300 detected boxes of Faster R-CNN \cite{ren2015faster} and drop each box's predicted class label. Next, we apply non-maximum suppression (NMS) over all the 300 boxes using a threshold of 0.5 to remove highly-overlapped ones. Basically, we trust the detected box locations (\ie, they do contain objects), but will recheck which of them belongs to the target object.  

To further reduce the number of candidate boxes, we sort the boxes by their initial detection confidence (in the descending order) and then remove the corresponding regions from the image \emph{in turn}\footnote{We crop out the corresponding image regions and replacing them by gray-color patches.}, every time followed by applying the image classifier to the resulting image. We stop this process until the classification confidence of the target class goes below a certain threshold. We then collect the removed box locations, which together have likely covered the target objects (high recall, but likely low precision), to be the candidate box pool for the next step.

\begin{figure*}[]
    \centerline{\includegraphics[width=0.78\linewidth]{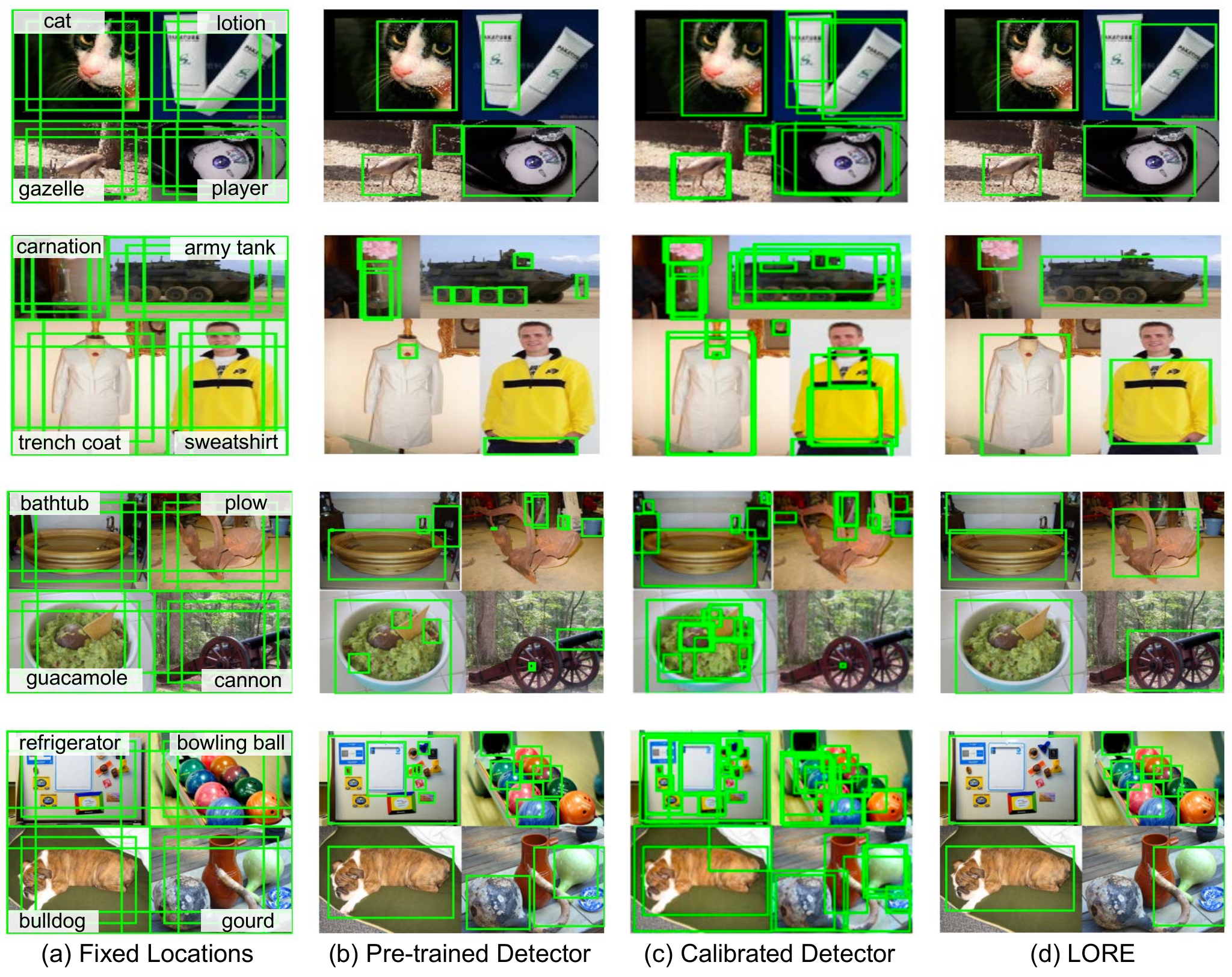}}
    \caption{\small \textbf{Box locations of different pseudo-label generation methods.} We show (a) fixed locations, (b) trust the pre-trained detector, (c) trust the calibrated detector, and (d) localization by region removals (LORE). The green boxes are the pseudo ground-truth locations found on each object-centric image alone before multiple images are stitched together. We can see that LORE accurately locates the target object in each sub-image while detection results are much noisy.
    Image class labels are listed on the corner of each sub-image in column (a).}
    \label{fig:suppl_psci}
\end{figure*}

\mypara{Localization by removal.}
To accurately identify which candidate truly belongs to the target class, we \emph{re-rank} the candidates by how much removing each boxed region \emph{alone} reduces the image classifier’s confidence on the target class. We then follow the descending order to remove these boxed regions \emph{in turn} until the classifier fail to predict the target class or the \emph{confidence reducing ratio}\footnote{We define the \emph{confidence reducing ratio} as the relative confidence drop on the target class label before and after removing boxes.} achieves a certain threshold. Finally, the bounding boxes of the removed regions are collected as the pseudo ground-truths for the image. 
More examples can be found in \autoref{fig:suppl_psci}.

\begin{table}[t]
\small
\tabcolsep 3.2pt
\renewcommand\arraystretch{1.0}
\centering
\caption{\small \textbf{Results with different pseudo-labels.} We use ImageNet-21K as the source of object-centric images and report the results of object detection on LVIS v0.5 val. \textbf{Detector}: object detector used for generating pseudo-label bounding boxes; \textbf{CL:} assign each box the image Class Label instead of the predicted class label.}
\begin{tabular}{l|cc|cccc}
 & Detector & CL & $\text{AP}^{b}$    &$\text{AP}_{r}^{b}$   & $\text{AP}_{c}^{b}$   & $\text{AP}_{f}^{b}$ \\ 
\shline 
Faster R-CNN$\star$ & -- & -- &  23.35&	12.98&	22.60&	28.42 \\ 
\hline
 \multirow{4}{*}{{\ourmethod}} & Pre-trained & {\color{gray}{\xmark}} & 23.04 & 13.93 & 21.51 & 28.14 \\
& Pre-trained & \cmark & 24.66 & 17.45 & \textbf{23.62} & {28.83}\\ 
& Calibrated & {\color{gray}{\xmark}} & 24.03 & 13.13 & 23.51 & \textbf{29.04}\\ 
 & Calibrated & \cmark & \textbf{24.93} & \textbf{19.31} & 23.51 & 28.95 \\
\hline
\end{tabular}\label{table:calib_detector}
\end{table}

\subsection{Discussion on fixed locations vs. LORE}
Both fixed locations and LORE use accurate image class labels.
Even though LORE gives more accurate object locations (see \autoref{fig:suppl_psci}) in pseudo-label generation, the resulting detector with fixed location is just slightly worse than that with LORE.
We attribute this small gap partially to two-stage fine-tuning, which adapts the detector back to accurately labeled scene-centric images. As shown in \autoref{table:suppl_fixed}, LORE notably surpasses fixed locations if we apply single-stage fine-tuning. 

\begin{table}[t]
\small
\tabcolsep 6pt
\renewcommand\arraystretch{1.0}
\centering
\caption{\small \textbf{Fixed locations vs. LORE.} We report object detection results on LVIS v0.5 val. \textbf{P-GT}: ways to generate pseudo-labels.}
\begin{tabular}{l|c|cccc}
         & P-GT & $\text{AP}^{b}$  & $\text{AP}_{r}^{b}$ & $\text{AP}_{c}^{b}$ & $\text{AP}_{f}^{b}$ \\
        \shline
        \multirow{2}{*}{Single-stage} & Fixed & 20.09 & 12.96 & 19.08 & 24.20 \\
         & LORE & 21.44 & 14.95 & 20.74 & 24.91 \\
        \hline
        \multirow{2}{*}{\ourmethod} & Fixed & 24.75 & 19.73 & 23.44 & 28.39 \\
         & LORE & 24.83 & {20.06} & 23.25 & 28.71 \\
         \hline
        \end{tabular}\label{table:suppl_fixed}
\end{table}

\subsection{Discussion on pseudo-label generation}
In this subsection, we discuss multiple ways for generating pseudo-labels in object-centric images. From the viewpoint of teacher models (\emph{i.e., } the pre-trained detector learned from a long-tailed distribution), we found that (1) the pre-trained detector is biased toward head classes, missing many accurate rare class predictions which have lower confidence scores; (2) detector calibration is useful to discover more bounding boxes for rare and common objects but can hardly correct wrong predicted labels. Our observations share the similar insights with a recent study~\cite{dave2021evaluating} on large-vocabulary object detection.

From the other viewpoint of fine-tuning with pseudo scene-centric images, we found that imputation using image class labels leads to a notable performance gain regardless of inaccurate box locations (\eg, fixed box locations). 
This is probably due to two reasons. First, dense boxes (like six fixed locations) can be treated as data augmentation for training the object detector. Second, our two-stage fine-tuning is beneficial in learning with noisy data, \emph{i.e.,} first on noisy pseudo scene-centric images and then on the clean labeled data from LVIS.

Other possibilities for pseudo-label generation include (1) iteratively improving the teacher detector by noisy student learning~\cite{xie2020self} and (2) calibrating the detector with more advanced approaches for class-imbalanced semi-supervised learning~\cite{wei2021crest}, etc.


\section{Additional Ablation on Image Mosaicking}
\label{suppl_mosaic}

\begin{figure}[t]
    \centerline{\includegraphics[width=1\linewidth]{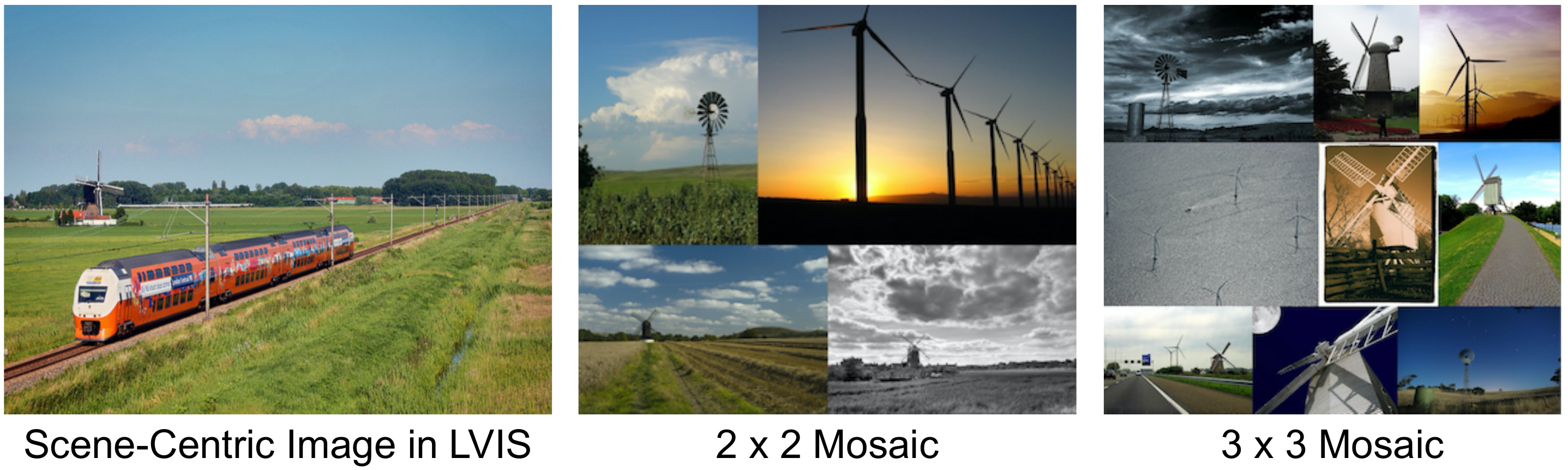}}
    \caption{\small \textbf{Different layouts of mosaics.} We show different types of mosaics from the same category (``windmill''). The $2\times2$ mosaic image (middle) and the real scene-centric image (left) in the LVIS dataset look alike in terms of appearance and structure while the $3\times3$ mosaic image (right) is much crowded.}
    \label{fig:suppl_mosaic}
\end{figure}

\emph{Does mosaicking more images help?}
In this section, we investigate the effect of different types of layouts for stitching object-centric images, \ie, $1\times1$ (which is the original object-centric image), $2\times2$ mosaic, and $3\times3$ mosaic. We evaluate them under the same experimental settings: we use ImageNet-21K as the source of object-centric images ($1,016$ classes) and stitch images from the same class and use the 6 fixed locations as pseudo ground-truths. \autoref{table:suppl_3_by_3} shows the comparison of object detection results on LVIS v0.5 dataset. We see that $2\times2$ and $3\times3$ mosaics perform similarly and both outperform the $1\times1$ OCI (on $\text{AP}^{b}$ and $\text{AP}_r^{b}$). An example with different layouts of $2\times2$ and $3\times3$ mosaics is shown in \autoref{fig:suppl_mosaic}.

\begin{table}[t]
\small
\tabcolsep 10pt
\renewcommand\arraystretch{1.0}
\centering
\caption{\small \textbf{Comparison of different types of mosaic images.} Here we use ImageNet-21K as the source of object-centric images and stitch images from the \emph{same} class and use the 6 fixed locations as pseudo ground-truths. $1\times1$ OCI means directly using the original object-centric images. Results are reported on LVIS v0.5 val. We can see that $2\times2$ mosaic gives better performance on all classes. The best result per column is in bold font.}
\begin{tabular}{l|cccc}
& $\text{AP}^{b}$    &$\text{AP}_{r}^{b}$   & $\text{AP}_{c}^{b}$   & $\text{AP}_{f}^{b}$ \\ 
\shline
Faster R-CNN$\star$ & 23.35&	12.98&	22.60&	28.42 \\ 
\hline
$1\times1$ OCI& 24.27 & 16.97 & \textbf{23.29} & \textbf{28.42} \\ 
$3\times3$ Mosaic & 24.29 & 18.14	& 23.13	& 28.21 \\
$2\times2$ Mosaic & \textbf{24.48}	&	\textbf{18.76} &	{23.26}& {28.29} \\
\hline
\end{tabular}\label{table:suppl_3_by_3}
\end{table}

\section{Further Analysis on Self-training}
\label{suppl_self_train}
We show detailed comparison results of self-training baseline in \autoref{table:self_train} to further demonstrate the effectiveness of the mosaicking and two-stage fine-tuning in our \ourmethod framework. We follow the self-training method with the normalization loss in \cite{zoph2020rethinking}.

\mypara{Mosaicking is also beneficial for self-training.}
We first study the vanilla self-training that directly learns object-centric (without mosaicking) and scene-centric images jointly. Specifically, we apply the pre-trained detector to generate pseudo-labels on the object-centric images (D). Next, the pre-trained detector is trained to jointly optimize the losses on human labels from LVIS and pseudo labels on object-centric images. We compare with and without image mosaic in \autoref{table:self_train}: image mosaicking improves $\text{AP}^{b}/\text{AP}^{b}_{r}$ from $22.00/14.04$ to $22.71/14.52$, demonstrating the effectiveness of mosaicking object-centric images to mitigate the domain discrepancy between two types of images. 

\mypara{Self-training vs. our two-stage fine-tuning.}
To further improve the performance of self-training, we apply 
``trusted the calibrated detector + image class labels'' (D$\ddagger$) as the pseudo-labeling method, which leads to a much higher detection accuracy than ``trusted the pre-trained detector'' (D) for our \ourmethod (cf. \autoref{table:main} of the main paper and the last row vs. the first row in \autoref{table:self_train}).
With this pseudo-labeling method, we see a notable gain against ``trust the pre-trained detector'' (D) for self-training. 

We further compare the self-training procedure that fine-tunes the detector simultaneously with object-centric and scene-centric images to our \ourmethod with two-stage fine-tuning (again in \autoref{table:self_train}).
\ourmethod outperforms self-training (with either D or D$\ddagger$) in most metrics, demonstrating the strength of two-stage fine-tuning which first learns with object-centric images and then scene-centric images. This two-stage pipeline is not only robust to noisy pseudo scene-centric data but also able to tie the detector to its final application domain with real scene-centric images.

\begin{table}[t]
\small
\tabcolsep 3.5pt
\renewcommand\arraystretch{1.0}
\centering
\caption{\small \textbf{Comparison to self-training.} \textbf{Mosaic:} \cmark means 2$\times$2 image mosaicking from different classes. \textbf{P-GT:} ways to generate pseudo-labels (\textbf{D}: trust the pre-trained \text{d}etector, \textbf{D$\ddagger$}: trust the calibrated detector and image class label). }
\begin{tabular}{l|cc|cccc}
 & Mosaic & P-GT & $\text{AP}^{b}$    &$\text{AP}_{r}^{b}$   & $\text{AP}_{c}^{b}$   & $\text{AP}_{f}^{b}$ \\ 
\shline 
Faster R-CNN$\star$ & -- & -- &  23.35&	12.98&	22.60&	28.42 \\ 
\hline
\multirow{3}{*}{{Self-training}} & {\color{gray}{\xmark}} & D & 22.00  & 14.04 & 20.41 & 27.18 \\ 
 & \cmark & D &  22.71 & 14.52 &	21.41 &	27.61 \\ 
 & \cmark & ~~D$\ddagger$ & 23.65  & 16.30 & 22.55 & 27.96 \\ 
\hline
\multirow{2}{*}{{\ourmethod}} & \cmark & D & 23.04 & 13.93 & 21.51 & 28.14 \\
&\cmark  & ~~D$\ddagger$ & 24.93 & 19.31 & {23.51} & {28.95}\\ 
\hline
\end{tabular}\label{table:self_train}
\vspace{-3mm}
\end{table}


\begin{figure*}[h]
    \centerline{\includegraphics[width=0.95\linewidth]{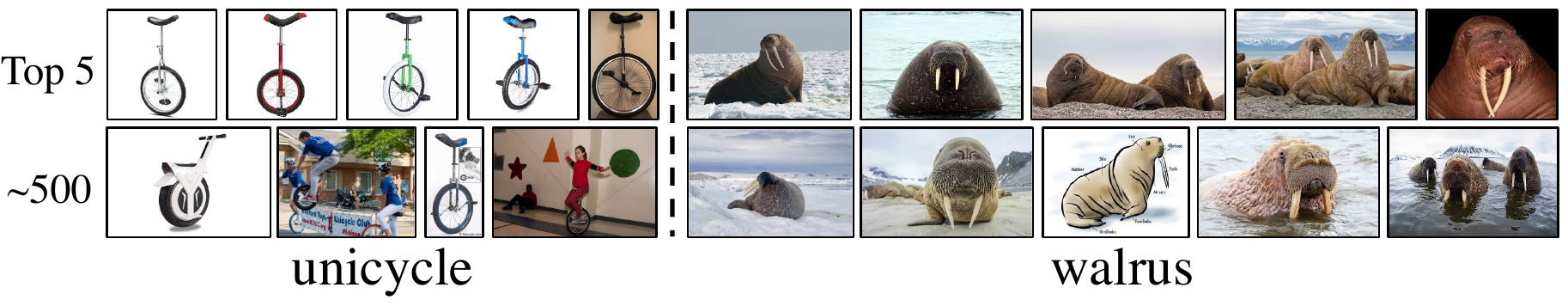}}
    \caption{\small \textbf{Google Images for the most rare classes in LVIS.} We show the top 5 retrieved images and images ranked around 500.}
    \label{fig:suppl_google}
\end{figure*}

\section{Data Quality of Object-Centric Images}
\label{suppl_oci}

Our main results are based on the ImageNet dataset~\cite{deng2009imagenet}. We included Google/Flickr images (\autoref{table:data} in the main text) mainly to analyze the effect of data quality and compare to \cite{ramanathan2020dlwl}. 
As shown in \autoref{fig:suppl_google}, most Google images searched by object names are object-centric, even for those not ranked on the top.
Following the experimental setup in \autoref{table:data} of the main paper, we further experiment with 500 Google images per class: $\text{AP}^{b}$ is improved from $24.45$ to $24.63$. 
For images that are less object-centric, LORE can give better pseudo-labels than the fixed heuristic; our two-stage fine-tuning is robust to noise.
Moreover, there are extensive works on de-noising web data that we can leverage to further improve our scalability and applicability.
That being said, we neither focus on web images/crowd-sourcing nor suggest that human efforts (\eg, ImageNet) are not needed. Our claim is that rare objects that are hard to collect from SCI are easier to collect from OCI, which opens up a new way to tackle long-tailed object detection.

\section{Comparison to Adversarial Training}
\label{suppl_adv_training}

We apply adversarial training (\eg, \cite{ganin2016domain}) to jointly train the detector with LVIS and pseudo scene-centric images.
Concretely, we train an additional domain classifier to differentiate the LVIS images and pseudo scene-centric images, and incorporate a gradient reversal layer (GRL) \cite{ganin2016domain} to minimize the discrepancy between their features to overcome the domain gap.
We show comparisons in \autoref{table:suppl_adv_training}. Adversarial training outperforms naive joint (\ie, single-stage) training, and  \ourmethod (with two-stage fine-tuning using each source) surpasses adversarial training.

\begin{table}[t]
\tabcolsep 6.5pt
\renewcommand\arraystretch{1.0}
\centering
\small
\caption{\small {\textbf{Comparison to adversarial training.} Results are reported on LVIS v0.5 validation set.}}
\begin{tabular}{l|c|cccc}
         & Mosaic & $\text{AP}^{b}$  & $\text{AP}_{r}^{b}$ & $\text{AP}_{c}^{b}$ & $\text{AP}_{f}^{b}$ \\
        \shline
        {Single-stage} & \cmark & 20.09 & 12.96 & 19.08 & 24.20 \\
        \hline
        \multirow{2}{*}{Adv. training}
         & \xmark & 20.92 &	11.42&	19.23&	26.85 \\
        & \cmark & 22.87 & 14.48 & 22.02 & 27.28 \\
        \hline
        \multirow{2}{*}{\ourmethod}  & \xmark & {24.27} & {16.97} & {23.29} & \textbf{28.42}  \\
        & \cmark & \textbf{24.75} & \textbf{19.73} & \textbf{23.44} & {28.39} \\
         \hline
        \end{tabular}\label{table:suppl_adv_training}
\end{table}

\section{Implementation Details of \ourmethodbf}
\label{suppl_impl}

\subsection{Details on object detection}
\label{supp_impl_detection}
As mentioned in \S~{\color{red}{6.1}} of the main paper, we use {Faster R-CNN} \cite{ren2015faster}
as our base detector and further extend the training process with another 90K iterations and select the checkpoint with the best $\text{AP}^{b}$ as {Faster R-CNN$\star$}. We use Faster R-CNN$\star$ as our main baseline to ensure that the improvement of \ourmethod does not simply come from training (\ie, fine-tuning) with more epochs.

For \ourmethod, we first fine-tune {Faster R-CNN$\star$} with pseudo scene-centric images, and then fine-tune it with the LVIS training set again. Both stages are trained end-to-end with stochastic gradient descent with all training losses in \autoref{eq:1} of the main paper, using a mini-batch size of $16$, momentum of $0.9$, weight decay of $10^{-4}$, and learning rate of $2\times 10^{-4}$. \emph{Unlike other long-tailed methods~\cite{tan2020equalization,tan2020equalizationv2,wang2020seesaw}\footnote{Both EQL(v2)~\cite{tan2020equalization,tan2020equalizationv2} and Seesaw loss~\cite{wang2020seesaw} introduce (multiple) additional hyper-parameters.}, there is no additional hyper-parameter in our framework.}

\subsection{Details on instance segmentation}
\label{suppl_inst}
\mypara{Background on instance segmentation.} 
We apply Mask R-CNN \cite{he2017mask}, which adopts the two-stage network architecture similar to Faster R-CNN \cite{ren2015faster}, with an identical first stage RPN. In the second stage, in addition to predicting the class label and box offset, Mask R-CNN further outputs a binary segmentation mask for each
proposal. 
Formally, during training, the entire Mask R-CNN is learned with four loss terms
\begin{align}
    \mathcal{L} = \mathcal{L}_\text{rpn} + \mathcal{L}_\text{cls} + \mathcal{L}_\text{reg} + \mathcal{L}_\text{mask},
    \label{eq:2}
\end{align}
where the RPN loss $\mathcal{L}_\text{rpn}$, classification loss $\mathcal{L}_\text{cls}$, and box regression loss $\mathcal{L}_\text{reg}$ are identical to those defined in \cite{ren2015faster}. The mask loss $\mathcal{L}_\text{mask}$ is learned via an average binary cross-entropy objective.

\mypara{Multi-stage training for instance segmentation.} 
We first train a Mask R-CNN using labeled scene-centric images from LVIS with instance segmentation annotations \cite{gupta2019lvis}. All the fours loss terms in \autoref{eq:2} are optimized. 

We then fine-tune the model using the pseudo scene-centric images that are generated from object-centric images. We use these images (only with box pseudo-labels) to fine-tune the model using $\mathcal{L}_\text{cls}$, $\mathcal{L}_\text{rpn}$, and $\mathcal{L}_\text{reg}$. In other words, we do not optimize $\mathcal{L}_\text{mask}$. Any network parameters that affect $\mathcal{L}_\text{cls}$, $\mathcal{L}_\text{rpn}$, and $\mathcal{L}_\text{reg}$, especially those in the backbone feature network (except
the batch-norm layers), can be updated.

After this stage, we fine-tune the whole network again with labeled scene-centric images from LVIS, using all the four loss terms in~\autoref{eq:2}.
The training procedure and other implementation details for instance segmentation are exactly the same as object detection in \S~\ref{supp_impl_detection}.
\section{Experimental Results on LVIS v0.5}
\label{suppl_v0.5}

Due to space limitations, we only compared with state-of-the-art methods in \autoref{table:sota_detection} and \autoref{table:sota_segmentation} of the main paper. In this section, we provide detailed comparisons with more previous works on LVIS v0.5. 

\mypara{Object detection on LVIS v0.5.}
There are not many papers reporting detection results on LVIS. In \autoref{table:stoa_detection_v0.5}, we further include EQL~\cite{tan2020equalization} and LST~\cite{hu2020learning}, together with BaGS~\cite{li2020overcoming} and TFA~\cite{wang2020frustratingly}, as the compared methods. \ourmethod outperforms all baselines except BaGS~\cite{li2020overcoming}.
We note that, BaGS is pre-trained on COCO~\cite{lin2014microsoft} while \ourmethod is initialized from ResNet-50 that is pre-trained on ImageNet-1K (ILSVRC). 
By using the COCO pre-trained backbone as the initialization, \ourmethod outperforms BaGS on nearly all metrics. 
Moreover, when combined with \cite{ren2020balanced}, \ourmethod can further boost the state-of-the-art performance.

\begin{table}[t]
\small
\tabcolsep 3.8pt
\renewcommand\arraystretch{1.0}
\centering
\caption{\small {{\textbf{Object detection on LVIS v0.5}. We use ImageNet + Google Images. MSCOCO: for pre-training. \cite{ren2020balanced}: balanced loss.} Within each column, {\color{red}{red}}/{\color{blue}{blue}} indicates the best/second best.}}
\begin{tabular}{r|cc|cccc}
\multicolumn{1}{c|}{} & MSCOCO &\cite{ren2020balanced} & $\text{AP}^{b}$ & $\text{AP}_{r}^{b}$   & $\text{AP}_{c}^{b}$   & $\text{AP}_{f}^{b}$  \\
\shline
RFS~\cite{gupta2019lvis} & & & 23.35 & 12.98 & 22.60 & 28.42 \\
EQL~\cite{tan2020equalization} & & & 23.30 & -- & -- &  -- \\
LST~\cite{hu2020learning} & & &22.60 & -- & -- &  -- \\
BaGS~\cite{li2020overcoming} & \cmark & & 25.96 & 17.65 & 25.75 & 29.54 \\
TFA~\cite{wang2020frustratingly} & &&24.40 & 16.90 & 24.30 & 27.70 \\
\hline
\multirow{4}{*}{\ourmethod} &&& 25.01 & {\color{blue}{20.25}} & 23.89 & 28.32 \\
&\cmark&& 26.28 & 17.37 & 26.13 & {\color{blue}{30.02}} \\
& &\cmark&  {\color{blue}{26.83}} & {\color{red}{21.00}} & {\color{blue}{26.31}} & 29.81\\
&\cmark&\cmark&  {\color{red}{{28.06}}} & 19.11 & {\color{red}{28.23}} & {\color{red}{31.41}}\\
\hline
\end{tabular}\label{table:stoa_detection_v0.5}
\end{table}

\begin{table}[t]
\tabcolsep 8.5pt
\renewcommand\arraystretch{1.0}
\centering
\small
\caption{\small {\textbf{Instance segmentation on LVIS v0.5.}
Our \ourmethod uses images from ImageNet and Google Images. +~\cite{ren2020balanced}: include the balanced loss in the second stage fine-tuning. Within each column, {\color{red}{red}}/{\color{blue}{blue}} indicates the best/second best.}}
\begin{tabular}{r|cccc}
\multicolumn{1}{c|}{} & $\text{AP}$ & $\text{AP}_{r}$   & $\text{AP}_{c}$   & $\text{AP}_{f}$  \\
\shline
RFS~\cite{gupta2019lvis} & 24.38 & 15.98 & 23.96 & 28.27 \\
EQL~\cite{tan2020equalization} & 22.80 & 11.30 & 24.70 & 25.10 \\
LST~\cite{hu2020learning}& 23.00 & -- & -- & -- \\
SimCal~\cite{wang2020devil} & 23.40 & 16.40 & 22.50 & 27.20 \\
Forest RCNN~\cite{wu2020forest} & 25.60 & 18.30 & 26.40 & 27.60 \\
BaGS~\cite{li2020overcoming} & 26.25 & 17.97 & 26.91 & 28.74 \\
BALMS~\cite{ren2020balanced} & 27.00 & 19.60 & {\color{red}{28.90}} & 27.50 \\
EQL v2~\cite{tan2020equalizationv2} & {\color{blue}{27.10}} & 18.60 & 27.60 & {\color{red}{29.90}}\\
\hline
\ourmethod & 26.26 & {\color{blue}{19.63}} & 26.60 & 28.49 \\
\ourmethod+~\cite{ren2020balanced} &  {\color{red}{27.86}} & {\color{red}{20.44}} & {\color{blue}{28.82}} & {\color{blue}{29.62}}\\
\hline
\end{tabular}\label{table:stoa_segmentation_v0.5}
\end{table}

\mypara{Instance segmentation on LVIS v0.5.}
The comparison results on LVIS 0.5 instance segmentation are presented in \autoref{table:stoa_segmentation_v0.5}, including the baseline models with RFS~\cite{gupta2019lvis} for re-sampling, EQL(v2)~\cite{tan2020equalization,tan2020equalizationv2} for re-weighting, LST~\cite{hu2020learning} for incremental learning, SimCal~\cite{wang2020devil} and BaGS~\cite{li2020overcoming} for de-coupled training, Forest R-CNN~\cite{wu2020forest} for hierarchy classification, and BALMS~\cite{ren2020balanced} for a balanced softmax loss. \ourmethod can perform on a par with or even better than the compared methods without any additional hyper-parameter tuning like in \cite{tan2020equalization,tan2020equalizationv2,wang2020seesaw}.
By combined with \cite{ren2020balanced}, \ourmethod achieves the stat-of-the-art performance of $27.86/20.44$ $\text{AP}/\text{AP}_{r}$, showing the compatibility of \ourmethod. We expect that \ourmethod could be further improved by incorporating other long-tailed learning strategies~\cite{li2020overcoming,wang2020frustratingly,ren2020balanced,tan20201st,wang2020seesaw}.
\section{Experimental Results on LVIS v1.0}

\label{suppl_lvis_v1}
\begin{table*}[th]
\small
\tabcolsep 15pt
\renewcommand\arraystretch{1.0}
\centering
\caption{\small \textbf{Number of overlapped classes in LVIS and ImageNet.} In LVIS and ImageNet, each category can be identifed by a unique WordNet synset ID. We match LVIS classes to ImageNet ones and show the number of the overlapped classes. 
Specifically, we show \# LVIS classes / \# overlapped to ImageNet-21K / \# overlapped to ImageNet-1K (ILSVRC).}
\begin{tabular}{c|c|ccc|c}
Version & Split & Frequent & Common & Rare & Overall \\
\shline
\multirow{2}{*}{v0.5} & Train & 315 / 253 / 85 & 461 / 387 / 96 & 454 / 385 / 71 & 1230 / 1025 / 252 \\
 & Val & 313 / 252 / 21 & 392 / 329 / 84 & 125 / 106 / 71 & 830 / 678 / 176 \\
 \hline
\multirow{2}{*}{v1.0} & Train & 405 / 331 / 87 & 461 / 390 / 96 & 337 / 277 / 64 & 1203 / 998 / 247 \\
 & Val & 405 / 331 / 87 & 452 / 382 / 92 & 178 / 144 / 37 & 1035 / 857 / 216 \\
 \hline
\end{tabular}\label{table:suppl_dataset_overlap}
\end{table*}

\begin{table}[h]
\small
\tabcolsep 8pt
\renewcommand\arraystretch{1.0}
\centering
\caption{\small \textbf{Statistics of LVIS v0.5 and v1.0 datasets.}}
\begin{tabular}{c|l|rrr}
 Version & \multicolumn{1}{c|}{Type} & Train & Val & Test \\
\shline
\multirow{3}{*}{v0.5} & \# Image & 57,263 & 5,000 & 19,761 \\
 & \# Class & 1,230 & 830 & - \\
 & \# Instance & 693,958 & 50,763 & - \\
\hline
\multirow{3}{*}{v1.0} & \# Image & 100,170 & 19,809 & 19,822 \\
 & \# Class & 1,203 & 1,035 & - \\
 & \# Instance & 1,270,141 & 244,707 & - \\
 \hline
\end{tabular}\label{table:suppl_stats_dataset}
\end{table}

\begin{table}[h]
\small
\tabcolsep 3.8pt
\renewcommand\arraystretch{1.2}
\centering
\caption{\small \textbf{Instance segmentation on LVIS v1.0.} We list multiple Mask R-CNN baselines whose accuracy are notably different due to differences in implementation, which may affect the accuracy of the corresponding proposed methods.
\ourmethod outperforms Mask R-CNN and many other methods on most of the metrics.} 
\begin{tabular}{c|r|cccc}
 Backbone & \multicolumn{1}{c|}{Method} & $\text{AP}$ &$\text{AP}_{r}$   & $\text{AP}_{c}$   & $\text{AP}_{f}$ \\ 
\shline

\multirow{9}{*}{R-50} & Mask RCNN~\cite{gupta2019lvis}$^{\dagger 1}$ &22.59 &		12.31 & 21.30 & 28.55 \\
& Mask RCNN~\cite{gupta2019lvis}$^{\star 2}$ & 23.70 &  13.50 & 22.80 & 29.30 \\
& Mask RCNN~\cite{gupta2019lvis}$^{\S 1}$ & 22.20 &  11.50 & 21.20 & 28.00 \\
& cRT~\cite{kang2019decoupling}$^{\S 1}$ & 22.10 &  11.90 & 20.20 & 29.00 \\
& BaGS~\cite{li2020overcoming}$^{\S 1}$ & 23.10 &  13.10 & 22.50 & 28.20 \\
& EQL v2~\cite{tan2020equalizationv2}$^{\S 1}$ & 23.70 &  14.90 & 22.80 & 28.60  \\
& EQL v2~\cite{tan2020equalizationv2}$^{\S 2}$ & 25.50 & 17.70 & 24.30 & 30.20 \\
& Seesaw~\cite{wang2020seesaw}$^{\star 2}$ & 26.40 &   19.60 &  26.10 &  29.80 \\ 
& \cellcolor[HTML]{EFEFEF}\ourmethod$^{\dagger 1}$ &\cellcolor[HTML]{EFEFEF}{24.49}  & \cellcolor[HTML]{EFEFEF}{18.30} &	\cellcolor[HTML]{EFEFEF}{23.00} &	\cellcolor[HTML]{EFEFEF}{28.87} \\ 

\hline

\multirow{6}{*}{R-101}& Mask RCNN~\cite{gupta2019lvis}$^{\dagger 1}$ & 24.82 &		15.18 & 23.71 & 30.31  \\
& Mask RCNN~\cite{gupta2019lvis}$^{\star 2}$ & 25.50 & 16.60 & 24.50 & 30.60 \\
& EQL~\cite{tan2020equalization}$^{\star 2}$ & 26.20 & 17.00 & 26.20 & 30.20 \\
& BaGS~\cite{li2020overcoming}$^{\star 2}$ & 25.80 & 16.50 & 25.70 & 30.10 \\
& Seesaw~\cite{wang2020seesaw}$^{\star 2}$ & 28.10 &  20.00 & 28.00 & 31.90 \\
& \cellcolor[HTML]{EFEFEF}{{\ourmethod}}$^{\dagger 1}$ &  \cellcolor[HTML]{EFEFEF}{26.77}   & \cellcolor[HTML]{EFEFEF}{20.79}  & \cellcolor[HTML]{EFEFEF}{25.76} & \cellcolor[HTML]{EFEFEF}{30.53} \\ 
\hline
\multirow{2}{*}{X-101} & Mask RCNN~\cite{gupta2019lvis}$^{\dagger 1}$ & 26.62 &		17.51 & 25.51 & 31.86 \\
& \cellcolor[HTML]{EFEFEF}{{\ourmethod}}$^{\dagger 1}$ &\cellcolor[HTML]{EFEFEF}{28.31}  & \cellcolor[HTML]{EFEFEF}{21.74} &	\cellcolor[HTML]{EFEFEF}{27.25} &	\cellcolor[HTML]{EFEFEF}{32.36} \\ 
\hline
\end{tabular}\label{table:lvis_v1_seg}
\begin{flushleft}
\vskip -.5em
$^\dagger$: Our implementations with RFS~\cite{gupta2019lvis}.\\
$^\star$: Results reported in \cite{wang2020seesaw}. All models trained with RFS~\cite{gupta2019lvis}. \\
$^\S$: Results reported in \cite{tan2020equalizationv2}. \\
$^1$: 1x schedule. $^2$: 2x schedule.
\end{flushleft}
\vskip -1em
\end{table}

\begin{table*}[h]
\small
\tabcolsep 9.5pt
\renewcommand\arraystretch{1.0}
\centering
\caption{\small \textbf{Comparisons of instance segmentation on LVIS v1.0.} \ourmethod achieves comparable improvements against the Mask R-CNN baseline. We note that, Seesaw loss~\cite{wang2020seesaw} uses a different implementation and training schedule (\ie, 2x). Thus, the results may not be directly comparable.}
\begin{tabular}{c|c|c|rrrr}
 Backbone & Schedule & Method & $\text{AP}$  &$\text{AP}_{r}$   & $\text{AP}_{c}$   & $\text{AP}_{f}$ \\ 
\shline
\multirow{4}{*}{R-50} & \multirow{2}{*}{2x} & Mask RCNN~\cite{gupta2019lvis}  & 23.70 &	13.50 &	22.80 &	29.30 \\
& & Seesaw~\cite{wang2020seesaw} &  {\color{teal}{\textbf{(+2.70)}}} 26.40 & {\color{teal}{\textbf{(+6.10)}}} 19.60 & {\color{teal}{\textbf{(+3.30)}}} 26.10 & {\color{teal}{\textbf{(+0.50)}}} 29.80 \\
\cline{2-7}
& \multirow{2}{*}{1x} & Mask RCNN~\cite{gupta2019lvis} & 22.59 &	12.31 & 21.30 & 28.55 \\
& & \cellcolor[HTML]{EFEFEF}\ourmethod &\cellcolor[HTML]{EFEFEF}{\color{teal}{\textbf{(+1.90)}}}  {24.49}  & \cellcolor[HTML]{EFEFEF}{\color{teal}{\textbf{(+5.99)}}} {18.30} &	\cellcolor[HTML]{EFEFEF}{\color{teal}{\textbf{(+1.70)}}} {23.00} &	\cellcolor[HTML]{EFEFEF}{\color{teal}{\textbf{(+0.32)}}} {28.87} \\ 
\hline
\multirow{4}{*}{R-101} & \multirow{2}{*}{2x} & Mask RCNN~\cite{gupta2019lvis}  & 25.50 & 16.60 & 24.50 & 30.60\\
& & Seesaw~\cite{wang2020seesaw} &  {\color{teal}{\textbf{(+2.60)}}} 28.10 & {\color{teal}{\textbf{(+3.40)}}} 20.00 & {\color{teal}{\textbf{(+3.50)}}} 28.00 & {\color{teal}{\textbf{(+1.30)}}} 31.90 \\
\cline{2-7}
 & \multirow{2}{*}{1x} & Mask RCNN~\cite{gupta2019lvis} &24.82 &	15.18 & 23.71 & 30.31  \\
& & \cellcolor[HTML]{EFEFEF}{\ourmethod} & \cellcolor[HTML]{EFEFEF}{\color{teal}{\textbf{(+1.95)}}} 26.77 & \cellcolor[HTML]{EFEFEF}{\color{teal}{\textbf{(+5.61)}}} 20.79 & \cellcolor[HTML]{EFEFEF}{\color{teal}{\textbf{(+2.05)}}} 25.76 & \cellcolor[HTML]{EFEFEF}{\color{teal}{\textbf{(+0.22)}}} 30.53 \\ 
\hline
\end{tabular}\label{table:lvis_v1_seg_compare}
\end{table*}

\subsection{Setup}
\mypara{Dataset statistics.}
We further evaluate \ourmethod on LVIS v1.0~\cite{gupta2019lvis}. The total dataset size has been expanded to $\sim$160K images and $\sim$2M instance annotations. The total number of categories has decreased slightly (from 1,230 to 1,203) due to a more stringent quality control. More specifically, LVIS v1.0 adds 52 new classes while drops 79 classes from LVIS v0.5. The validation set has been expanded from 5K images to 20K images. \autoref{table:suppl_stats_dataset} gives a summary of the statistics of the two versions of LVIS dataset. 
We follow the experimental setups of LVIS v0.5 to use category synset ID~\cite{miller1995wordnet} to search for the corresponding classes in ImageNet-21K dataset~\cite{russakovsky2015imagenet}. In total, we collect $753,700$ object-centric images. \autoref{table:suppl_dataset_overlap} shows the detailed statistics of the number of overlapped classes in those datasets. We also search 100 images for each class via Google Images.

\mypara{Our settings.}
For instance segmentation, we use {Mask R-CNN}~\cite{he2017mask} with instance segmentation annotations. The training scheme is the same as that for Faster R-CNN in object detection.
Specifically, we follow the default training configurations in \cite{wu2019detectron2} with 1x schedule\footnote{EQL v2~\cite{tan2020equalizationv2} and Seesaw loss~\cite{wang2020seesaw} use another implementation from~\cite{mmdetection}, which uses 2x schedule for training the models on LVIS v1.0.}.

For the \ourmethod training (cf. \S~\ref{suppl_inst}), we first fine-tune the baseline Mask R-CNN for 90K iterations with pseudo scene-centric images using only box annotations. Our pseudo scene-centric images are synthesized with $2\times2$ mosaic from random classes of ImageNet-21K and Google images. 
We use the boxes with 6 fixed locations as pseudo ground-truths.
After that, We end-to-end fine-tune the entire model for another 90K iterations using the LVIS training set with all four losses. The network parameters of the mask head are initialized by the baseline Mask RCNN model. Both two fine-tuning steps are trained with stochastic gradient descent with a mini-batch size of 16, momentum of 0.9, weight decay of $10^{-4}$, and learning rate of $2\times10^{-4}$.

\subsection{Instance segmentation on LVIS v1.0}
\autoref{table:lvis_v1_seg} shows detailed results on instance segmentation. We mainly compare with Mask R-CNN and two recent papers~\cite{tan2020equalizationv2,wang2020seesaw}, which reported instance segmentation results and re-implemented some other methods on LVIS v1.0.
We evaluate \ourmethod with three different backbone models: ResNet-50~\cite{he2016deep}, ResNet-101~\cite{he2016deep}, and ResNeXt-101~\cite{xie2017aggregated}: \ourmethod consistently outperforms the Mask R-CNN baseline especially for rare classes.

We note that, EQL v2~\cite{tan2020equalizationv2} and Seesaw loss~\cite{wang2020seesaw} were implemented by a different framework~\cite{mmdetection} and reported results with a stronger 2x training schedule. \emph{Thus, the accuracy gap between different methods may be partially affected by these factors.} This can be seen by comparing the three Mask R-CNN results with ResNet-50 and the two Mask R-CNN results with ResNet-101: there is a notable difference in their accuracy. Specifically, the ones reported by \cite{wang2020seesaw} have a much higher accuracy.

With the same ResNet-50 backbone and 1x schedule, \ourmethod achieves $24.49/18.30$ $\text{AP}/\text{AP}_{r}$, better than EQL v2~\cite{tan2020equalizationv2} ($23.70/14.90$), BaGS~\cite{li2020overcoming}, and cRT~\cite{kang2019decoupling}. With the ResNet-101 backbone, \ourmethod with 1x schedule achieves $26.54$ AP, outperforming both EQL~\cite{tan2020equalization} (2x schedule, $26.20$ AP) and BaGS~\cite{li2020overcoming} (2x schedule, $25.80$ AP). We also show a detailed comparison to Seesaw loss~\cite{wang2020seesaw} in \autoref{table:lvis_v1_seg_compare}. \ourmethod demonstrates a comparable performance gain against the Mask R-CNN baseline.
\section{Qualitative Results}
\label{suppl_qual}

\begin{figure*}[t]
    \centerline{\includegraphics[width=0.8\linewidth]{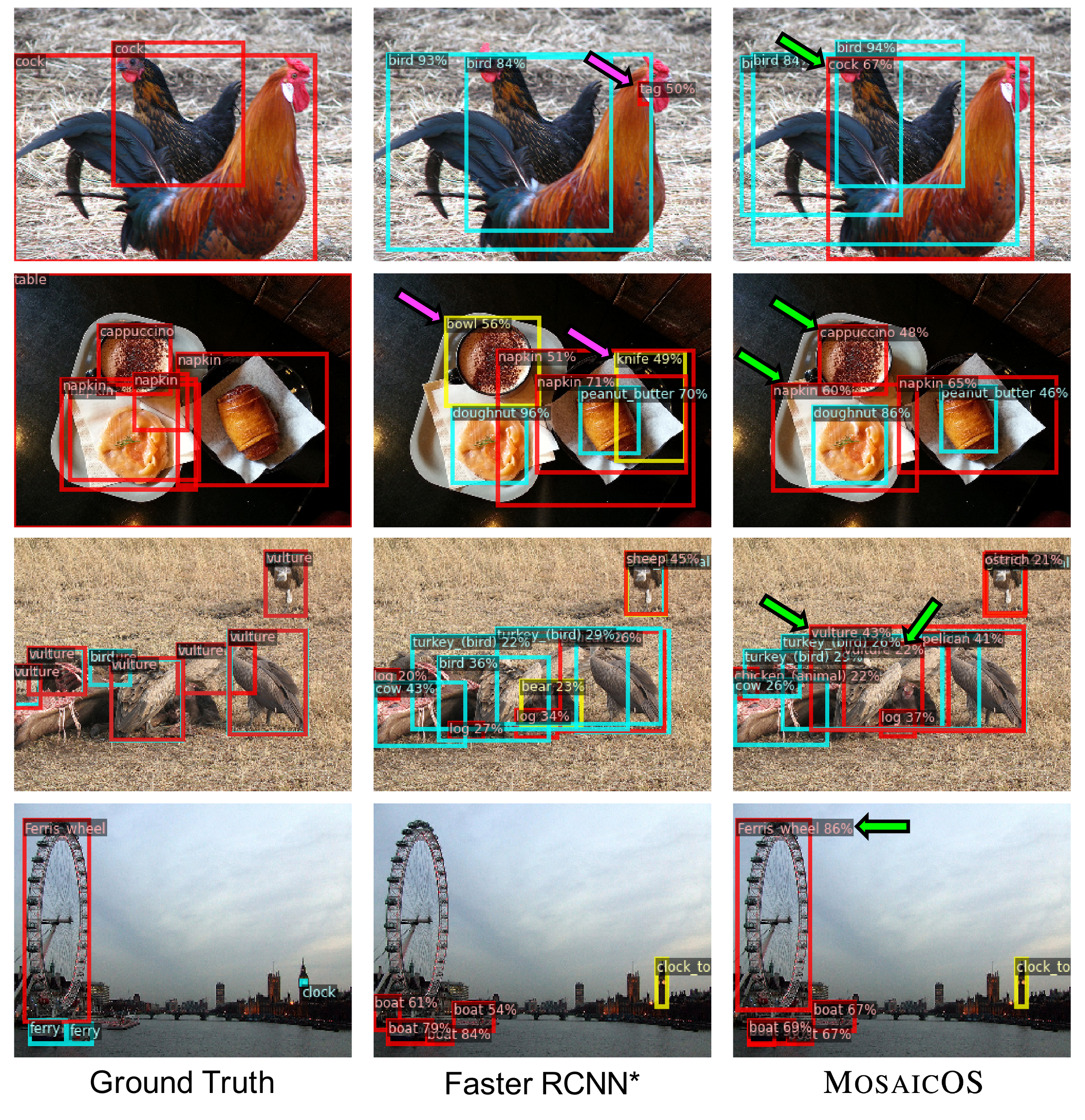}}
    \caption{\small \textbf{Qualitative results on object detection.} Our approach can detect the rare objects missed by the baseline detector (\eg, \emph{cock}, \emph{cappuccino}, \emph{ferris wheel}) and correct the labels that were wrongly classified to frequent categories (\eg, \emph{bear}, \emph{knife}, \emph{bowl}). We superimpose {\color{green}{\textbf{green}}} arrows to show where we did right while the baseline did wrong ({\color{magenta}{\textbf{magenta}}}). {\color{yellow}{\textbf{Yellow}}}/{\color{cyan}{\textbf{Cyan}}}/{\color{red}{\textbf{Red}}} boxes indicate frequent/common/rare (predicted) class labels.}
    \label{fig:suppl_qual1}
\end{figure*}

\begin{figure*}[t]
    \centerline{\includegraphics[width=0.8\linewidth]{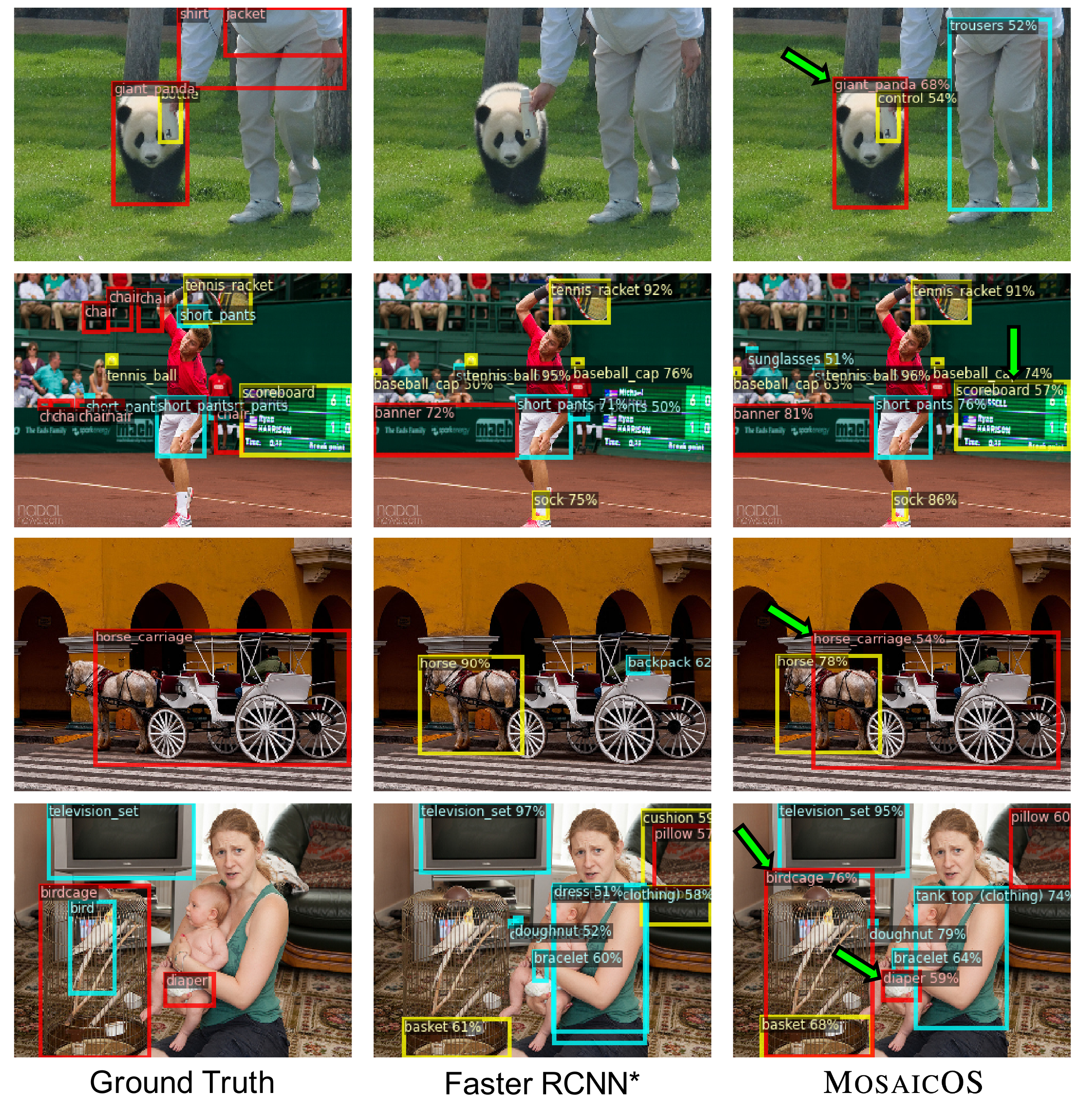}}
    \caption{\small \textbf{Additional qualitative results on object detection.}  We superimpose {\color{green}{\textbf{green}}} arrows to show that our approach can detect the objects missed  by the baseline detector (\eg, \emph{gaint panda}, \emph{scoreboard}, \emph{horse carriage}, \emph{birdcage}, \emph{diaper}). {\color{yellow}{\textbf{Yellow}}}/{\color{cyan}{\textbf{Cyan}}}/{\color{red}{\textbf{Red}}} boxes indicate frequent/common/rare (predicted) class labels.}
    \label{fig:suppl_qual2}
\end{figure*}

We show qualitative results on LVIS v0.5 object detection in \autoref{fig:suppl_qual1} and \autoref{fig:suppl_qual2}. We compare the ground truth, the results of the baseline and of our method. 

We observe that our method can accurately recognize more objects from rare categories that may be overlooked by the baseline detector. For example, as shown in \autoref{fig:suppl_qual2}, \ourmethod correctly detects giant panda, scoreboard, horse carriage, and diaper. They are all rare classes and the baseline detector fails to make any correct detection (\ie, localization and classification) on them. Moreover, the results demonstrate that \ourmethod is able to correct the prediction labels that were wrongly classified to frequent classes without sacrificing the detection performance on common and frequent classes. As shown in the second row of \autoref{fig:suppl_qual1}, the baseline detector wrongly predicts frequent class labels like bowl and knife with high confidence score, while \ourmethod suppresses them and successfully predicts rare classes napkin and cappuccino. 

One characteristic of LVIS is that the objects may not be exhaustively annotated in each image. We find that \ourmethod still detects those objects which are not labeled as the ground truths. In the second and third row of \autoref{fig:suppl_qual2}, the predictions on banner and horse are obviously correct while LVIS doesn't have annotations on them. 

\end{document}